\documentclass[11pt,a4paper]{article}

\usepackage[T1]{fontenc}
\usepackage[utf8]{inputenc}
\usepackage[a4paper,top=2.4cm,bottom=2.9cm,left=1.9cm,right=1.9cm]{geometry}
\usepackage{mathptmx}                 
\usepackage{graphicx}
\usepackage{amsmath}
\usepackage{amssymb}
\usepackage{array}
\usepackage{tabularx}
\usepackage{multirow}
\usepackage{setspace}
\usepackage{enumitem}
\usepackage{needspace}
\usepackage{lastpage}
\usepackage{fancyhdr}
\usepackage[font={bf},labelsep=period,justification=justified,
            singlelinecheck=false,skip=4pt]{caption}
\usepackage[numbers,sort&compress]{natbib}
\usepackage[hidelinks,breaklinks=true]{hyperref}
\makeatletter
\AtBeginDocument{\let\NAT@space\relax}
\makeatother

\graphicspath{{figures/}}

\newcolumntype{L}[1]{>{\raggedright\arraybackslash\hsize=#1\hsize}X}
\newcolumntype{C}[1]{>{\centering\arraybackslash\hsize=#1\hsize}X}
\newcolumntype{J}[1]{>{\hsize=#1\hsize}X}

\newcommand{\HeadingOne}[1]{%
  \par\addvspace{12pt}\needspace{3\baselineskip}%
  {\fontsize{14}{17}\selectfont\bfseries\raggedright #1\par}\addvspace{3pt}}
\newcommand{\HeadingTwo}[1]{%
  \par\addvspace{10pt}%
  {\fontsize{12}{15}\selectfont\bfseries\raggedright #1\par}\addvspace{2pt}}
\newcommand{\HeadingThree}[1]{%
  \par\addvspace{8pt}%
  {\normalsize\bfseries\raggedright #1\par}\addvspace{2pt}}

\newenvironment{wordbullets}
  {\begin{itemize}[leftmargin=0.8cm,labelsep=0.25cm,topsep=4pt,
                   itemsep=4pt,parsep=0pt,label=\textbullet]}
  {\end{itemize}}

\begin{document}

\begin{center}
{\fontsize{16}{19}\selectfont\bfseries
Design-to-Plan: A Large Language Model-Based Multi-Agent Framework for
Manufacturing Process Planning from 3D CAD Models and 2D Engineering Drawings\par}

\vspace{12pt}

{\normalsize\bfseries
Muhammad Tayyab Khan\textsuperscript{\,a,\,c*}, Lequn Chen \textsuperscript{b*},
Wenhe Feng\textsuperscript{\,a}, Seung Ki Moon \textsuperscript{c*}\par}

\vspace{8pt}

\begin{spacing}{1.5}
{\fontsize{9}{13}\selectfont
\textsuperscript{a\,}Singapore Institute of Manufacturing Technology (SIMTech),
Agency for Science, Technology and Research (A*STAR), 5 CleanTech Loop,
\#01-01 CleanTech Two Block B, Singapore 636732, Republic of Singapore\par
\textsuperscript{b\,}Advanced Remanufacturing and Technology Centre (ARTC),
Agency for Science, Technology and Research (A*STAR), 3 CleanTech Loop,
\#01-01 CleanTech Two, Singapore 637143, Republic of Singapore\par
\textsuperscript{c}\textsuperscript{\,}School of Mechanical and Aerospace
Engineering, Nanyang Technological University, 639798, Singapore\par
\textsuperscript{*} Corresponding authors:
\href{mailto:khan0022@e.ntu.edu.sg}{khan0022@e.ntu.edu.sg} (M.T. Khan),
\href{mailto:chen1470@e.ntu.edu.sg}{chen1470@e.ntu.edu.sg} (L. Chen),
\href{mailto:skmoon@ntu.edu.sg}{skmoon@ntu.edu.sg} (S.K. Moon)\par}
\end{spacing}
\end{center}

\HeadingOne{Abstract}

Manufacturing process planning requires transforming heterogeneous design information into coherent and executable manufacturing decisions. However, existing computational approaches typically focus on isolated subtasks, such as manufacturing feature recognition, drawing interpretation, or tool selection, and therefore struggle to support the complete reasoning chain from original design artifacts to final manufacturing process plans. This limitation becomes particularly critical when planning must jointly interpret 3D CAD models, 2D engineering drawings, material information, and domain-specific manufacturing rules. To address the gap, this paper presents \textbf{Design-to-Plan}, a large language model (LLM)-based multi-agent framework for end-to-end manufacturing process planning. The proposed framework introduces an agentic planning architecture in which an orchestrator agent coordinates specialized agents across the full workflow, including 3D manufacturing feature recognition, 2D drawing analysis, 2D-3D context fusion, manufacturing knowledge retrieval, process sequencing, tool selection, and report generation. Rather than using LLMs as standalone text generators, the framework deploys the LLMs as interactive reasoning agents that communicate with deterministic modules, external knowledge sources, and one another to produce consistent and traceable planning decisions. A hybrid deterministic-agentic design is adopted, where deterministic modules and specialized agents extract structured information from CAD and drawing inputs, while LLM agents perform context-aware reasoning, retrieve manufacturing rules, resolve conflicts, and generate planning outputs. The framework is evaluated using 300 benchmark cases across the three downstream ReAct-enabled agents, together with separate evaluations of CAD feature recognition, drawing analysis, and 2D-3D context fusion. The parallel
architecture achieves a 100\% success rate across evaluated downstream agents,
Tool F1 scores of 95.9\%--97.6\%, 90\% source detection accuracy in conflict
analysis, and a 60\%--68\% reduction in token usage for key downstream planning
tasks. These results demonstrate that structured LLM-based multi-agent
coordination can bridge heterogeneous design representations and manufacturing
knowledge, enabling scalable, efficient, and traceable design-to-plan
automation.


\HeadingOne{1. Introduction}

Manufacturing process planning requires the integration of geometric analysis,
engineering specifications, domain knowledge, and sequencing logic.
Traditionally, this task is performed by experienced engineers who
simultaneously consider part geometry, tolerances, material constraints,
machining processes, and tooling economics. However, increasing component
complexity and shorter product development cycles are driving the need for
automation. Although computer-aided process planning (CAPP) systems have been
studied for decades, conventional variant and generative approaches rely on
predefined templates that cannot adequately accommodate the diversity of
materials, geometries, and process requirements in modern manufacturing \cite{chang1985introduction,xu2017machinetool40}.

Over the past decade, machine learning (ML) methods have achieved significant
progress in individual subtasks within the process planning pipeline. Automatic
feature recognition (AFR) approaches based on graph neural networks demonstrate
high accuracy in classifying machining features from STEP files \cite{lambourne2021brepnet,colligan2022hierarchicalcadnet,khan2024automaticfeature}.
Vision-based models for engineering drawings have achieved strong performance in
detecting geometric dimensioning and tolerancing (GD\&T) callouts and
dimensional annotations using deep learning (DL)-based detection and document
understanding techniques \cite{redmon2016yolo,kim2022donut}. In addition, process planning rule systems have
been developed using expert systems and structured knowledge bases (KBs) \cite{leokumar2019knowledgebased,xiao2023knowledgegraph}.
Despite these advances, such solutions remain largely isolated. For instance,
AFR modules typically generate geometric outputs that are not directly linked to
downstream tasks, limiting their ability to support integrated knowledge
retrieval, process recommendation, and manufacturability analysis.

Recent advances in large language models (LLMs) and agentic AI frameworks
provide a promising pathway to overcome this fragmentation. Beyond text
generation, LLMs demonstrate emerging capabilities in structured reasoning \cite{wei2022chainofthought},
tool use \cite{schick2023toolformer,yao2023react}, embodied perception and action \cite{driess2023palme}, and cross-domain
knowledge synthesis \cite{achiam2023gpt4}. When embedded within multi-agent systems (MAS), these
capabilities can be organized into coordinated workflows in which specialized
agents exchange information, invoke external tools, retrieve domain knowledge,
and collectively solve complex decision-making problems \cite{wu2024autogen,bytez2023hugginggpt,park2023generativeagents}. Such
properties are highly relevant to manufacturing process planning, where
decisions must be made across heterogeneous design representations,
interdependent planning stages, and domain-specific constraints. However,
despite the rapid growth of LLM-powered MAS in other domains \cite{li2024surveyllmmas,guo2024llmmasurvey}, their
applications in manufacturing remain at an early stage. Existing studies on LLMs
for process planning and design understanding are largely conceptual or
task-specific, with limited implementation of inter-agent communication,
workflow-level coordination, and end-to-end evaluation \cite{wu2024autogen,makatura2023howcanllm}. As a result, the
potential of LLM-based multi-agent coordination for transforming fragmented
manufacturing planning tasks into an integrated design-to-plan workflow remains
underexplored.

To address these limitations, this paper presents Design-to-Plan, which, to the
best of the authors' knowledge, is among the first fully implemented agentic
frameworks that connect heterogeneous design artifacts with executable
manufacturing process planning through multi-agent reasoning. The proposed
system consists of six specialized agents that exchange structured information
asynchronously to perform feature extraction, context fusion, knowledge
retrieval, process sequencing, tool selection, and report generation. Unlike
prior work, which is largely conceptual or limited to isolated subtasks, the
proposed framework provides a coordinated design-to-plan workflow with
structured data exchange, traceable intermediate outputs, and systematic
end-to-end evaluation across diverse input conditions.

The architecture adopts a hybrid deterministic-agentic design in which
well-defined perception tasks are handled by reliable deterministic modules,
while LLM-based agents perform context-aware reasoning, knowledge retrieval,
conflict resolution, and manufacturing decision synthesis. Rather than using
LLMs as standalone text generators, the proposed framework deploys the LLMs as
interactive reasoning agents. These agents communicate with deterministic
services, external manufacturing knowledge sources, and other specialized agents
to generate consistent, traceable, and executable planning outputs. Structured
inter-agent coordination enables state persistence, thread tracking, and
asynchronous information flow across the complete design-to-plan workflow
without shifting the focus toward low-level implementation details.

The main contributions of this work are as follows: (1) A fully implemented
agentic design-to-plan framework for bridging the gap between design and
manufacturing, transforming heterogeneous design artifacts into executable
manufacturing process plans through coordinated multi-agent reasoning. (2) An
interactive LLM-based agentic planning architecture in which specialized agents
reason over design context, manufacturing knowledge, process constraints, and
tool requirements through structured communication rather than isolated text
generation. (3) A hybrid deterministic--agentic workflow that combines reliable
extraction of design information with flexible reasoning over ambiguous,
incomplete, and potentially conflicting manufacturing inputs. (4) A systematic
end-to-end benchmark evaluation across diverse input-complexity scenarios,
comparing sequential and parallel ReAct architectures and assessing agent
behavior under heterogeneous and conflicting knowledge sources.

\HeadingOne{2. Literature Review}

\HeadingTwo{2.1 Multi-Agent Systems in Manufacturing}

MAS have been studied in manufacturing for several decades, primarily in
flexible manufacturing control, shopfloor scheduling, and distributed production
management \cite{chan2002multiagentagile,acm_introduction_multiagent,springer_roadmap_agent}. Early holonic and agent-based architectures demonstrated
the effectiveness of distributed control, where agents representing machines,
jobs, or resources coordinate task allocation \cite{valckenaers2015designunexpected}. These systems improve
adaptability and fault tolerance compared with centralized approaches.

More recently, MAS have been extended to process planning and decision support.
Shen et al. \cite{shen1999agentbased} proposed collaborative frameworks in which agents from
different engineering domains exchange constraints and negotiate solutions.
Digital twin-based systems further enable real-time coordination between
physical machines and virtual models \cite{tao2019digitaltwin}, while STEP-NC-based MAS support
process planning for prismatic components \cite{nassehi2006stepnc}. Despite these advances, most
systems rely on predefined rules and static knowledge representations, limiting
their ability to generalize to new geometries, incomplete specifications, or
design information expressed across multiple sources.

LLM-based MAS introduce a new paradigm for agent coordination. Recent surveys
\cite{li2024surveyllmmas,guo2024llmmasurvey,he2025llmmasse} highlight rapid growth across domains while noting limited adoption
in manufacturing. Emerging systems include LLM-enhanced modular production
platforms \cite{xia2023towardsautonomous}, intelligent shopfloor management systems \cite{zhao2026llmshopfloor}, and embodied
multi-agent production frameworks \cite{liu2026llmembodied}. However, these applications focus
primarily on scheduling or control rather than end-to-end manufacturing process
planning from design artifacts. A key limitation remains the lack of flexible,
knowledge-intensive reasoning for handling non-standard geometries, ambiguous
annotations, and missing material data. Such capabilities have only recently
become feasible with LLM-based agents.

\HeadingTwo{2.2 LLMs for Manufacturing Intelligence}

The use of LLMs in manufacturing is rapidly expanding. Recent surveys report
applications in intelligent manufacturing \cite{zhang2025surveyllmintelligentmfg}, mechanics and product design
\cite{mustapha2025surveymechanics}, process planning and quality control \cite{li2026llmmanufacturing}, and next-generation
manufacturing systems \cite{ma2025leveragingllm}. LLMs such as GPT-4, LLaMA, and Claude demonstrate
broad technical knowledge derived from large-scale training corpora, while
prompting techniques such as chain-of-thought (CoT) and few-shot learning
improve reasoning on engineering tasks \cite{wei2022chainofthought,brown2020fewshot}.

The ReAct framework combines reasoning with tool use, enabling models to
iteratively query external resources such as databases and computational tools
\cite{yao2023react}. This addresses a key limitation of standalone LLMs, which rely on static
knowledge and lack grounding in validated external sources or task-specific
data. Toolformer further demonstrates autonomous tool usage, and modern
function-calling interfaces enable practical deployment of such systems \cite{schick2023toolformer}.

In manufacturing, prior work has explored LLMs for mechanical design
understanding \cite{makatura2023howcanllm}, tolerancing, material selection, and manufacturability
analysis \cite{ma2025leveragingllm,li2024llm4cad,zhang2026llmcadsurvey,daareyni2025genaicad}. However, most approaches rely on single LLMs operating
without structured tool use, multi-source knowledge integration, or coordinated
agent interaction. This limitation is critical, as manufacturing decisions must
be grounded in verified information from heterogeneous sources rather than
general-purpose knowledge alone.

Frameworks such as AutoGen \cite{wu2024autogen}, MetaGPT \cite{hong2023metagpt}, and CrewAI \cite{crewai_documentation} provide
infrastructure for building MAS with role specialization and structured
communication. However, these frameworks are designed for general-purpose
applications and require substantial adaptation for manufacturing, including
domain-specific tools, structured knowledge sources, and traceable decision
workflows. As a result, existing systems do not provide fully implemented,
end-to-end solutions spanning perception, 2D-3D context fusion, knowledge
retrieval, process planning, and tool selection.

\HeadingTwo{2.3 Feature Extraction from CAD Models and Drawings}

AFR from 3D CAD models has evolved from rule-based methods \cite{babic2008reviewafr} to DL
approaches. Early methods such as FeatureNet \cite{zhang2018featurenet} used 3D convolutional neural
networks (CNNs) on voxelized models. More recent graph neural network approaches
operating on boundary representation (B-Rep) topology improve performance by
capturing geometric relationships \cite{jayaraman2021uvnet}. BRepNet \cite{lambourne2021brepnet} introduced topological
message passing, while hierarchical GCNN models further enhanced feature
representation \cite{colligan2022hierarchicalcadnet}. Recent approaches such as AAGNet \cite{wu2024aagnet}, BRepGAT \cite{lee2023brepgat}, and
BrepMFR \cite{zhang2024brepmfr} extend feature recognition to segmentation and domain adaptation.
Vision-language models (VLMs) provide an alternative for complex geometries,
although with higher computational cost and reduced geometric precision
\cite{khan2025vlmfeaturerecognition,picard2025concepttomanufacturing}.

Engineering drawings provide complementary information not encoded in CAD
models, including tolerances, GD\&T annotations, surface finish requirements,
and manufacturing notes. Extraction methods have progressed from optical
character recognition (OCR)-based approaches to DL-based detection and document
understanding models \cite{khan2026drawingstodecisions}. Layout-aware architectures such as LayoutLMv3 \cite{huang2022layoutlmv3}
and DocLLM \cite{wang2024docllm} enable structured parsing of complex documents. Models such as
Donut \cite{kim2022donut} support end-to-end OCR-free extraction, while YOLO-based detectors
localize annotation regions. Hybrid pipelines that combine detection and
structured parsing improve robustness by decomposing the problem into manageable
subtasks \cite{khan2025multistagehybrid}.

A fundamental limitation across this literature is the lack of integration with
downstream manufacturing planning tasks. Feature recognition outputs are
typically evaluated independently and are not connected to knowledge retrieval,
process planning, or tool selection. Similarly, drawing annotations are rarely
mapped to corresponding 3D features. However, manufacturing decisions depend on
this feature-level linkage; for example, tolerance values and surface finish
requirements are only meaningful when associated with specific CAD features such
as holes, pockets, slots, or machined faces. This gap highlights the need for
integrated approaches that connect CAD-derived geometry and drawing-derived
specifications with downstream manufacturing intelligence.

\HeadingTwo{2.4 Process Planning and Tool Selection}

CAPP approaches range from variant methods, which reuse existing plans, to
generative methods that synthesize new plans from first principles \cite{chang1985introduction,xu2017machinetool40}.
Knowledge-based systems and knowledge graphs have been widely used to represent
manufacturing rules and constraints \cite{leokumar2019knowledgebased,xiao2023knowledgegraph}. However, these systems require
extensive manual knowledge engineering and struggle with inputs that fall
outside predefined templates.

Recent research explores LLM-based approaches to process planning. For example,
CAPP-GPT generates process plans directly from part descriptions \cite{azab2024cappgpt}, while
other studies investigate multimodal LLMs for CAD understanding and
manufacturing applications \cite{daareyni2025genaicad}. Tool selection has also been extensively
studied, including cutting parameter optimization and material-specific
recommendations \cite{groover_fundamentals}. Deterministic lookup-based approaches remain common in
industry due to their reliability and auditability. Despite these advances, most
systems address individual tasks in isolation and lack the ability to reason over
fused CAD--drawing context or reconcile incomplete and conflicting manufacturing
information, which is common in real-world manufacturing scenarios.

\HeadingTwo{2.5 Research Gaps}

The reviewed literature indicates that the key limitation of existing
manufacturing intelligence methods is not the absence of individual
capabilities, but the lack of integration among them. Despite progress in
multi-agent systems, CAD feature recognition, engineering drawing
interpretation, knowledge retrieval, and CAPP, most existing methods remain
focused on isolated stages of the process planning workflow rather than on a
connected design-to-manufacturing reasoning chain.

A fundamental gap lies in the limited correlation between 3D CAD information and
2D engineering drawing information. Existing 3D feature recognition methods can
identify manufacturing features from CAD models, while drawing interpretation
methods can extract dimensions, GD\&T annotations, surface finish requirements,
material information, and manufacturing notes. However, these outputs are
typically handled separately, and their integration still depends largely on
manual interpretation by manufacturing engineers. This is problematic because
downstream planning tasks, including knowledge retrieval, process sequencing,
and tool selection, require feature-specific manufacturing context. For example,
a tolerance or surface finish requirement extracted from a drawing is not
directly useful for planning unless it is linked to the corresponding hole,
pocket, slot, or other 3D manufacturing feature.

Furthermore, this missing 2D-3D linkage creates subsequent limitations in
manufacturing decision-making. Conventional CAPP and rule-based systems
generally assume that required planning inputs are already available in a
structured form, such as predefined feature lists, known materials, complete
dimensions, tolerance requirements, and process constraints. In practical design
scenarios, however, this information is distributed across CAD models, drawings,
annotations, notes, and manufacturing knowledge sources, and may be incomplete,
ambiguous, or expressed using non-standard terminology. More importantly,
downstream planning also requires reasoning across heterogeneous sources, where
rules, material constraints, and process recommendations may be incomplete or
inconsistent.

Therefore, the central research gap addressed in this work is the absence of a
fully implemented agentic framework that bridges heterogeneous design
representations and downstream manufacturing planning. This work targets that
gap by coordinating 2D-3D context fusion, knowledge retrieval, process
sequencing, and tool selection through structured inter-agent communication and
tool-grounded interactive reasoning.

\HeadingOne{3. Framework Architecture and Methodology}

\HeadingTwo{3.1 Overall Architecture}

The proposed framework adopts a three-tier architecture that executes the
end-to-end manufacturing process planning pipeline, as illustrated in Fig. 1. Tier 1, the Orchestrator Agent, coordinates the overall workflow, manages user
sessions, and integrates human-in-the-loop (HITL) interaction without performing
domain-specific computation. Tier 2, the Helper Agent layer, comprises six
domain-specific agents: Feature Extraction, Context Fusion, Knowledge Retrieval
(KR), Process Sequence (PS), Tool Selection (TS), and Report Generation. Tier 3,
the Utility Services layer, provides stateless ML services for 3D CAD feature
recognition and 2D drawing analysis, which are accessed exclusively by the
Feature Extraction Agent. This architectural design enables independent scaling
and efficient allocation of computational resources.

The proposed framework follows an interaction-driven coordination model. Rather
than invoking downstream stages through a rigid sequential pipeline, agents
communicate using structured message-passing protocols. This design supports
asynchronous execution, fault tolerance through retry mechanisms, and
independent scalability of framework components.

\begin{figure}[!htbp]
\centering
\includegraphics[width=1\textwidth,height=0.95\textheight,keepaspectratio]{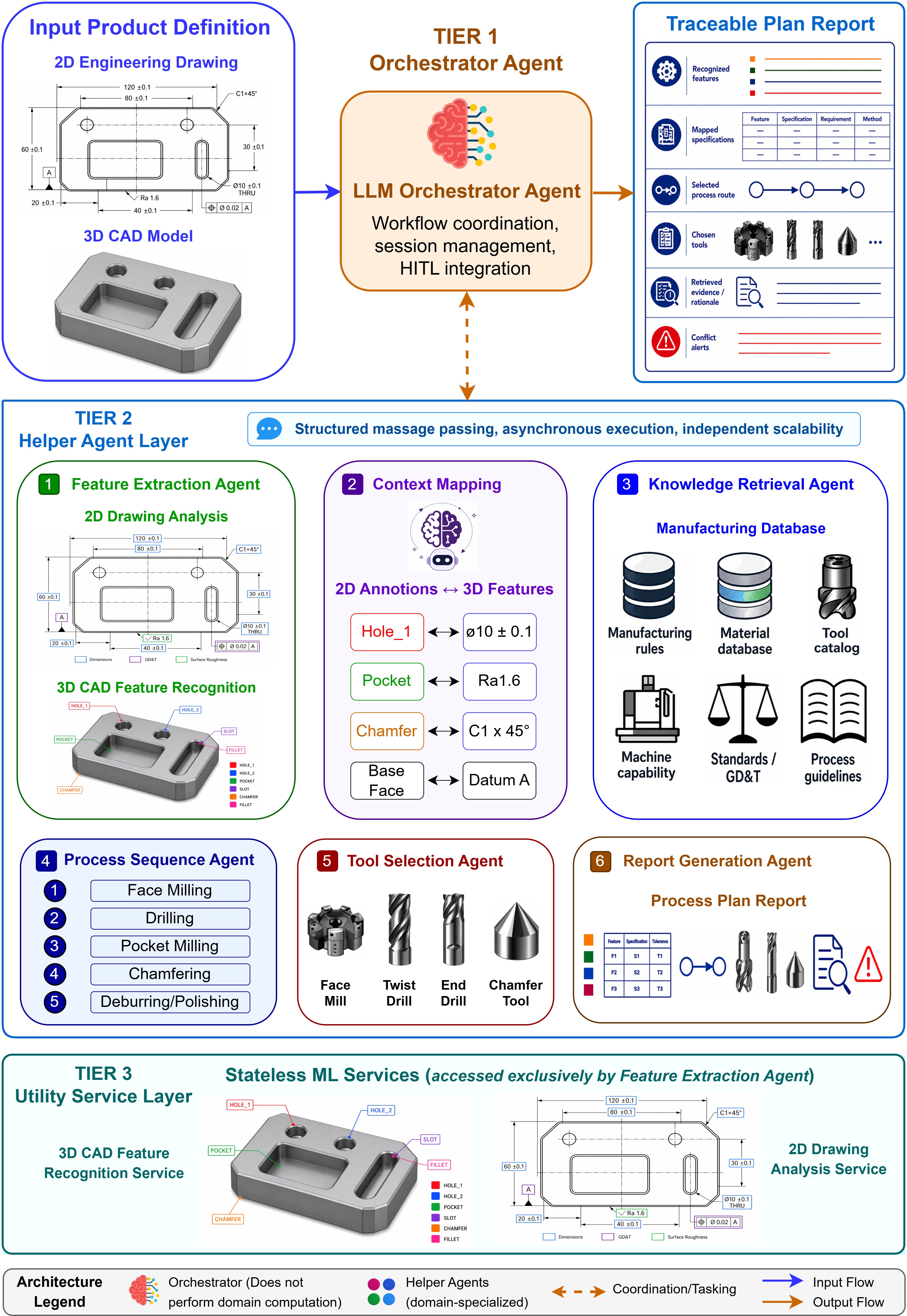}
\caption{Three-tier architecture of the proposed Design-to-Plan framework,
comprising orchestration and session management, six specialized planning agents,
and stateless ML services for CAD feature recognition and drawing analysis.}
\end{figure}

A web-based interface provides the user interaction layer. Engineers upload CAD
models in STEP format and engineering drawings in PDF, PNG, or JPG formats. The
system operates asynchronously, allowing users to monitor progress and review
intermediate results without blocking the interface. HITL capabilities enable
users to correct extracted features, annotations, or mappings before downstream
planning continues.

The Orchestrator Agent serves as the system entry point and session manager. It
handles file uploads, initializes session contexts, dispatches tasks to
specialized agents, monitors execution status, and aggregates intermediate
results. Importantly, the Orchestrator does not perform domain-specific
manufacturing reasoning; instead, it coordinates the workflow, maintains session
state, and routes structured information between agents. This design separates
workflow management from manufacturing decision-making, allowing each
specialized agent to focus on its assigned task while preserving traceability
across the design-to-plan process.

Perceptual processing is handled by two specialized agents. The Feature
Extraction Agent executes a stateful workflow using a directed-graph structure,
coordinating the extraction of geometric features from 3D CAD models and
manufacturing specifications from 2D engineering drawings, and then aggregating
these outputs into structured intermediate results. The Context Fusion Agent
links drawing-based annotations, including dimensions, GD\&T, surface finish
requirements, and notes, to the corresponding 3D CAD features. User corrections
from the HITL review stage are incorporated before producing a unified feature
representation for downstream planning.

Analytical processing is performed by four planning-oriented agents. The KR
Agent retrieves relevant manufacturing knowledge from heterogeneous sources and
resolves incomplete or conflicting information through tool-grounded reasoning.
The PS Agent generates and validates machining operation sequences based on the
fused design context and retrieved constraints. The TS Agent selects suitable
tools and computes machining parameters for the planned operations. The Report
Generation Agent consolidates outputs from all stages into a structured
manufacturing report. Together, these agents transform fused design information
into executable manufacturing planning outputs.

A manufacturing planning session proceeds through multiple coordinated stages.
After file upload, the Orchestrator Agent initializes the session and dispatches
the Feature Extraction task, where CAD and drawing inputs are processed
concurrently. The user then reviews intermediate outputs and provides corrections
if necessary. The Orchestrator subsequently triggers Context Fusion, followed by
knowledge retrieval, process sequencing, tool selection, and report generation.
Throughout this workflow, agents exchange structured messages asynchronously,
enabling non-blocking execution, traceable intermediate outputs, and modular
coordination across the design-to-plan pipeline. The overall workflow is shown
in Fig. 2.

\begin{figure}[!htbp]
\centering
\includegraphics[width=0.962\textwidth,height=0.84\textheight,keepaspectratio]{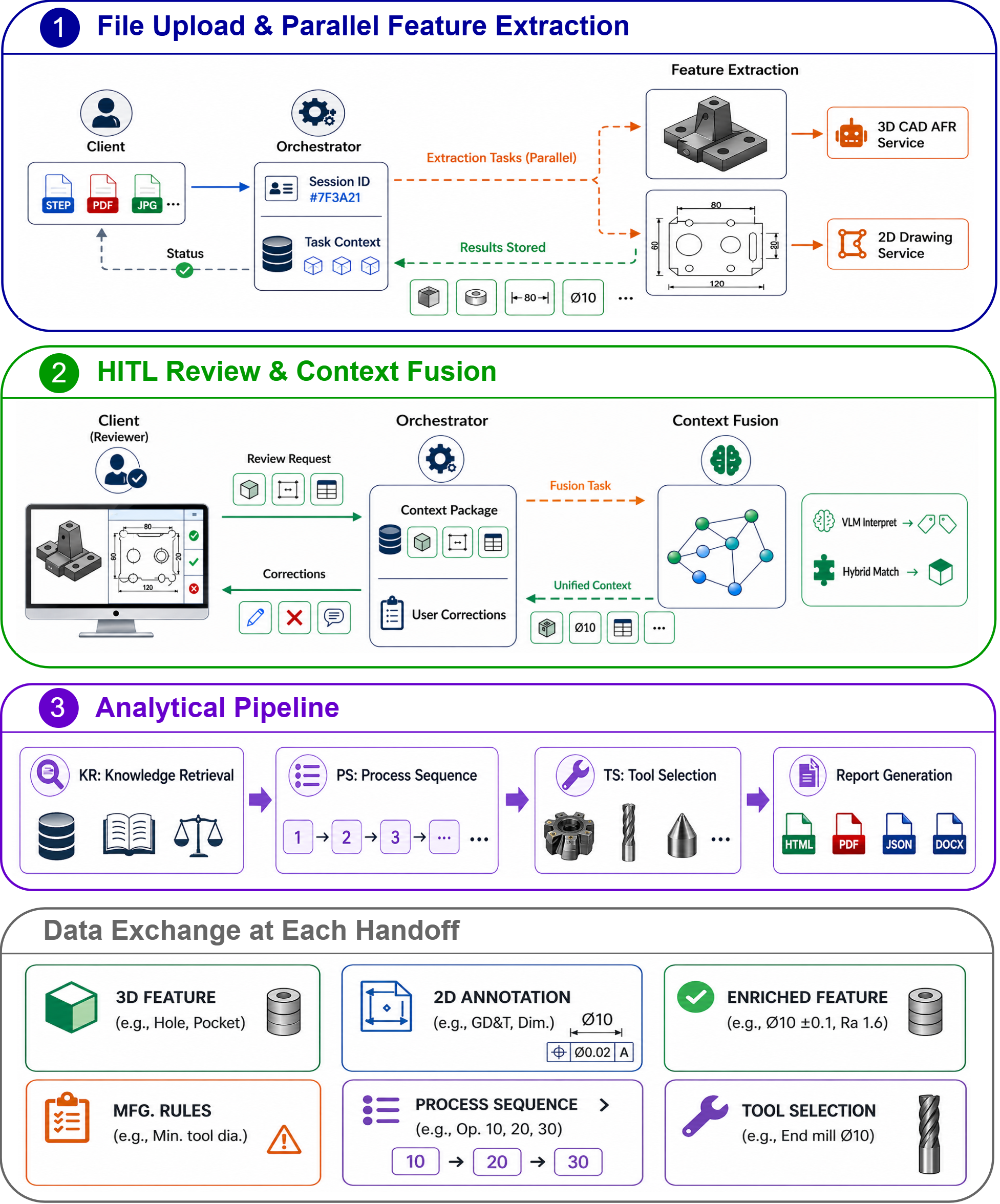}
\caption{End-to-end workflow of the proposed system, including file upload,
parallel feature extraction, HITL review, context fusion, and downstream
planning using progressively enriched design representations.}
\end{figure}

\HeadingTwo{3.2 Feature Extraction Agent}

The Feature Extraction Agent coordinates the extraction of geometric features
from 3D CAD models and manufacturing specifications from 2D drawings. It
receives an extraction request from the Orchestrator Agent, invokes specialized
ML services, and returns aggregated structured results. The agent operates as a
stateful directed-graph workflow, enabling coordinated execution,
synchronization, and resumption across multiple processing steps.

A key design feature of this agent is its parallel execution structure, as
illustrated in Fig. 3. After receiving an extraction request, the agent
simultaneously dispatches two independent processing branches. The 3D branch
invokes the CAD feature recognition service to identify manufacturing features
and extract geometric attributes, while the 2D branch invokes the drawing
analysis service to extract manufacturing specifications, including dimensions,
GD\&T annotations, surface roughness requirements, notes, and title-block
information. These branches run asynchronously, allowing CAD and drawing
analysis to proceed concurrently rather than sequentially. Once both branches
are completed, the directed-graph workflow enters a synchronization node that
waits for the outputs from both branches. Only after both results are available
does the agent resume execution and merge them into a structured intermediate
representation. This wait-and-resume mechanism ensures that downstream agents
receive a complete design context containing both CAD-derived manufacturing
features and drawing-derived engineering specifications before context fusion
begins.

\HeadingThree{3.2.1 CAD Feature Recognition}

CAD feature recognition uses a hierarchical GCNN to classify manufacturing
features from STEP files, as shown in the 3D CAD feature recognition branch of
Fig. 3. Full model details are provided in \cite{khan2024automaticfeature}; here, we summarize its
integration within the system.

Each STEP file is converted into a B-Rep representation, where faces are nodes
and adjacency relationships define graph connectivity. Each face is encoded with
geometric features, such as surface area, centroid, and surface type, while
three adjacency matrices represent convex, concave, and other edge
relationships. A finer facet-level representation captures local geometric
detail, with facet adjacency encoding spatial relationships.

The GCNN operates in two stages. First, face-level features are processed using
edge-typed graph convolutions:

\begin{equation}
H' = E_{1}HW_{1} + E_{2}HW_{2} + E_{3}HW_{3} + HW_{I} + b
\end{equation}

where $H \in R^{N_{f} \times d}$ is the face feature matrix, $E_{1}$, $E_{2}$,
and $E_{3}$ are adjacency matrices, $W_{1}$, $W_{2}$, and $W_{3}$ are learnable
weights, $W_{I}$ is the self-connection term, and $b$ is a bias vector.

These operations propagate information across different geometric relationships,
capturing topological patterns such as concave cavities and convex protrusions.
The resulting embeddings are projected into the facet space, where a
second-stage graph convolution is applied:

\begin{equation}
H' = A_{2}HW + HW_{I} + b
\end{equation}

where $A_{2}$ is the facet adjacency matrix. Face and facet representations are
linked through transfer operations, enabling joint modeling of global topology
and local geometry. The model outputs probabilities over 36 manufacturing
feature classes, including holes, pockets, slots, and chamfers. It also extracts
geometric attributes such as dimensions, orientations, and parameters required
for downstream planning.

\HeadingThree{3.2.2 Drawing Analysis Pipeline}

The drawing analysis pipeline uses a three-stage hybrid framework to extract
structured manufacturing specifications, as shown in the 2D drawing analysis
branch of Fig. 3. Full details are provided in \cite{khan2025multistagehybrid}.

\begin{wordbullets}
\item \textbf{Stage 1 (Layout Detection):} Major layout regions, including
views, annotations, and metadata, are detected using YOLO-based models,
separating structured and unstructured content.

\item \textbf{Stage 2 (Annotation Localization):} Annotation elements are
localized using oriented bounding boxes, enabling detection of GD\&T callouts,
dimensions, and surface roughness symbols at arbitrary orientations. Surface
roughness detection remains challenging due to limited training data; missed
detections are partially mitigated through HITL review.

\item \textbf{Stage 3 (Structured Parsing):} Detected annotations are converted
into structured representations using category-specific schemas. GD\&T
annotations include symbols, tolerances, and datum references, while dimensional
annotations include nominal values, tolerance limits, and directionality.
Textual regions, such as notes and title blocks, are processed using a language
model to handle variability in format and terminology.
\end{wordbullets}

\begin{figure}[!htbp]
\centering
\includegraphics[width=\textwidth,height=0.84\textheight,keepaspectratio]{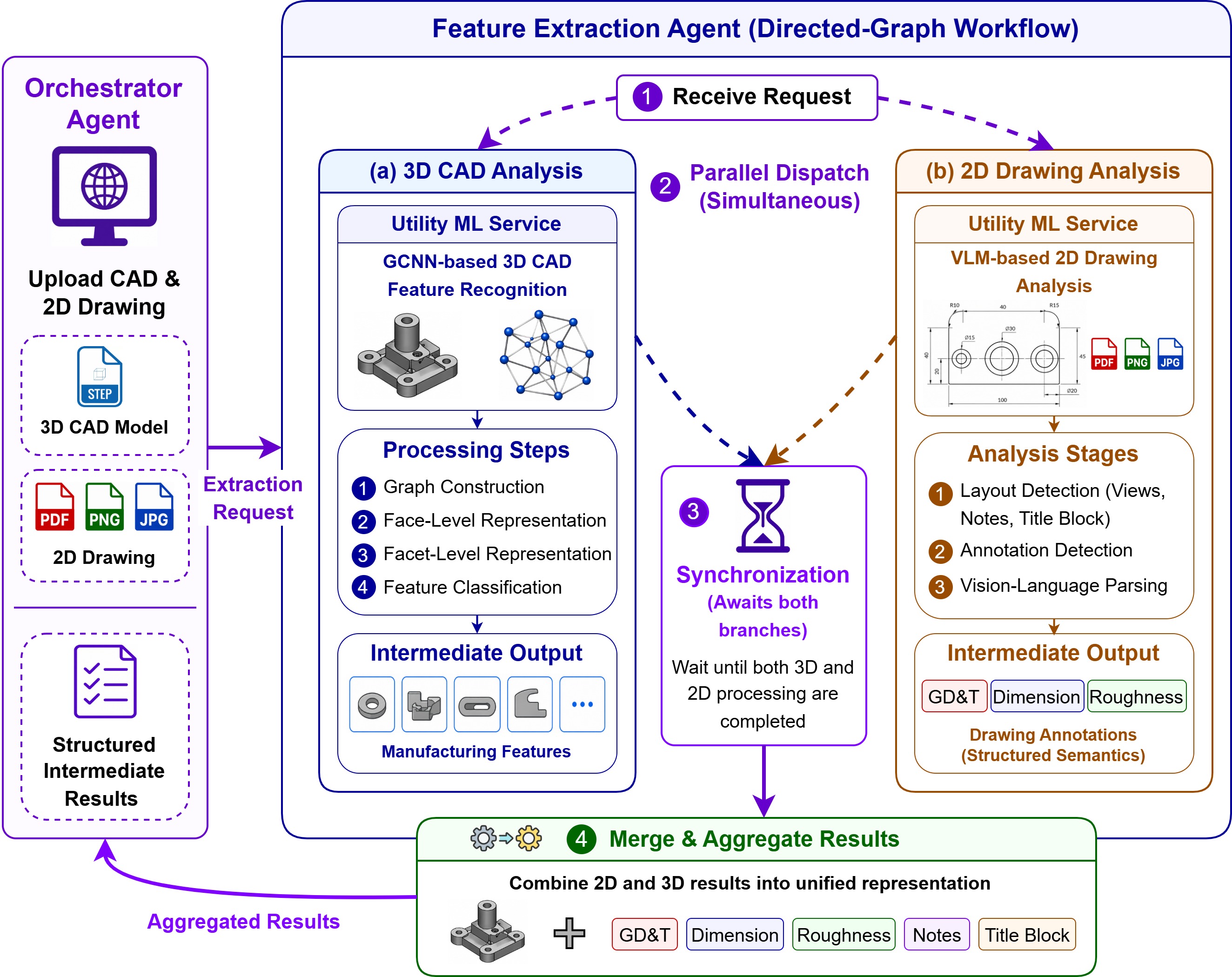}
\caption{Feature Extraction Agent workflow, showing parallel 3D CAD feature
recognition and 2D drawing analysis followed by synchronization and structured
output aggregation for context fusion.}
\end{figure}

\HeadingTwo{3.3 Context Fusion Agent}

The Context Fusion Agent is a central component of the proposed design-to-plan
workflow because it bridges the semantic gap between design representation and
manufacturing reasoning. The Feature Extraction Agent produces two separate
outputs: 3D CAD-derived manufacturing features and 2D drawing-derived
engineering specifications. However, these outputs are not directly actionable
unless they are correlated at the feature level. For example, a tolerance, GD\&T
callout, or surface finish requirement extracted from a drawing must be linked
to the specific hole, pocket, slot, chamfer, or face to which it applies before
it can support knowledge retrieval, process sequencing, or tool selection.

To address this requirement, the Context Fusion Agent generates a unified
feature representation by linking each 3D CAD feature with its corresponding 2D
drawing specifications, including dimensions, tolerances, GD\&T callouts,
surface finish requirements, notes, and other relevant manufacturing
information. The agent follows a two-stage workflow, as illustrated in Fig. 4.
First, semantic interpretation enriches each 2D annotation with manufacturing
meaning, such as target feature type, spatial context, and functional intent.
Second, hybrid matching uses this enriched annotation context together with 3D
feature attributes, including feature type, size, and location, to identify the
most plausible 2D-3D correspondences. User corrections from the HITL review
stage are incorporated before final mapping, ensuring that expert feedback can
override uncertain or incorrect automatic associations.

The resulting enriched 3D feature representation serves as the primary input for
downstream analytical agents. By converting isolated CAD features and drawing
annotations into feature-specific manufacturing context, the Context Fusion
Agent enables subsequent agents to retrieve relevant manufacturing rules,
generate valid process sequences, and select appropriate tools based on complete
and traceable design information.

\begin{figure}[!htbp]
\centering
\includegraphics[width=\textwidth,height=0.84\textheight,keepaspectratio]{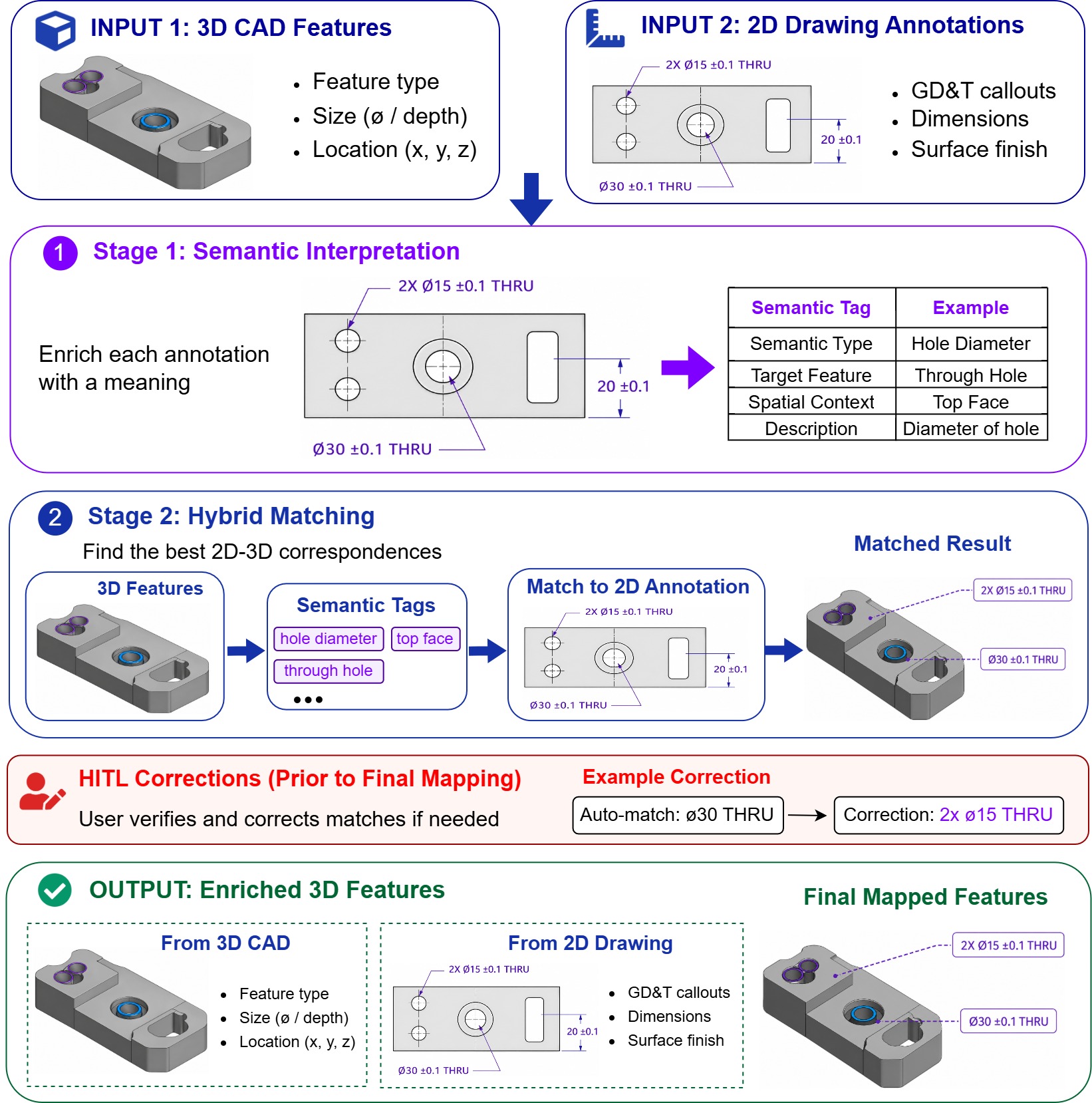}
\caption{Context fusion workflow for 2D-3D mapping, combining semantic
interpretation, hybrid feature matching, reasoning-based disambiguation, and
HITL correction to generate validated feature-level mappings.}
\end{figure}

\HeadingThree{3.3.1 Semantic Interpretation}

A key challenge in correlating 2D annotations with 3D features is ambiguity in
dimensional values. Multiple features may share identical dimensions while
representing different feature types; for example, a hole diameter and a pocket
width may have the same numerical value. To address this, each annotation is
enriched with semantic information using a VLM. The enriched representation
includes: (i) a semantic type describing functional meaning, such as hole
diameter or pocket depth; (ii) a descriptive interpretation capturing feature
grouping and design intent; (iii) an associated feature category indicating the
likely 3D feature type; and (iv) spatial context describing the approximate
location on the part. Manufacturing domain knowledge, including feature-type
vocabulary and spatial reasoning heuristics, is explicitly encoded in the VLM
prompt rather than assumed from general-purpose vision-language knowledge. As
shown in Fig. 5, the prompt defines the agent role, motivates semantic
disambiguation, specifies the required output fields for each annotation, and
encodes manufacturing heuristics for spatial reasoning and pattern recognition.
This prompt-guided enrichment prevents false matches based only on numerical
similarity and provides the semantic context required for robust 2D-3D feature
correlation.

\begin{figure}[!htbp]
\centering
\includegraphics[width=0.843\textwidth,height=0.84\textheight,keepaspectratio]{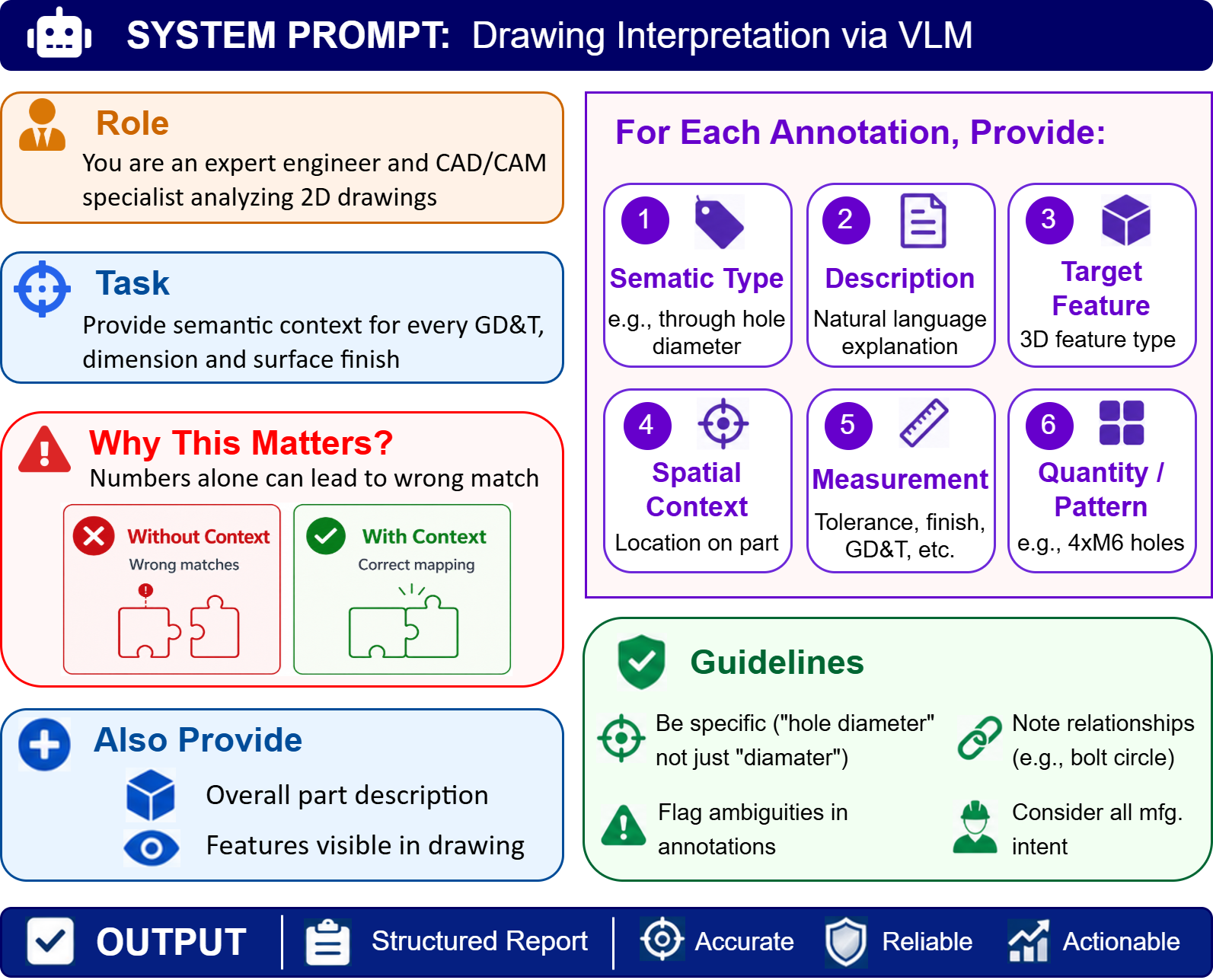}
\caption{VLM prompt design for semantic enrichment, defining the agent role,
output schema, and manufacturing heuristics for interpreting drawing annotations
and spatial context.}
\end{figure}

\HeadingThree{3.3.2 Hybrid 2D-3D Mapping}

The mapping stage correlates enriched 2D annotations with 3D features using a
hybrid scoring strategy. For each candidate pairing, a composite score is
computed:

\begin{equation}
S_{map} = w_{t} \cdot S_{type} + w_{d} \cdot S_{dim} + w_{s} \cdot S_{spatial}
\end{equation}

where $S_{type}$ denotes feature-type compatibility, $S_{dim}$ denotes
dimensional agreement, and $S_{spatial}$ denotes spatial consistency derived
from semantic interpretation. The weighting is adaptive based on data
availability. When 3D dimensional information is available, dimensional
agreement dominates the score. Otherwise, semantic and spatial cues are weighted
more heavily. This adaptive formulation improves robustness under incomplete or
ambiguous inputs. Additional adjustments are applied based on symbolic
consistency. For example, diameter symbols reinforce mappings to cylindrical
features, while radius indicators favor curved geometries. A near-tie filter
retains candidates close to the highest score, preventing premature elimination
of valid alternatives.

The mapping operates in two modes. In deterministic mode, high-confidence
matches are accepted directly. In reasoning-based mode, ambiguous cases are
resolved by evaluating candidate features using semantic descriptions and
geometric characteristics. Each mapping is assigned a confidence level:
high-confidence mappings are accepted automatically, medium-confidence mappings
are retained with caution, and low-confidence mappings trigger human review. The
system supports one-to-many relationships, allowing a single 3D feature to be
associated with multiple annotations. User corrections are incorporated prior to
finalization, with corrected values overriding automatic extraction. The final
output consists of enriched feature objects containing geometric attributes,
associated specifications, mapping confidence, and mapping method.

\HeadingTwo{3.4 Knowledge Retrieval Agent}

The Knowledge Retrieval (KR) Agent is the most analytically complex component in
Tier 2, the Helper Agent layer introduced in the three-tier architecture in
Fig. 1. It queries a multi-source manufacturing knowledge base (KB), detects and
resolves inconsistencies across sources, and returns structured constraints,
rules, and process recommendations relevant to the input features. The agent
operates using a ReAct loop within a stateful directed-graph framework, enabling
explicit tracking of intermediate reasoning steps for analysis and replay. The
agent supports three operational modes: deterministic retrieval, sequential
ReAct with a single reasoning agent, and parallel ReAct with multiple
specialized sub-agents operating concurrently. The overall architecture is shown
in Fig. 6.

\begin{figure}[!htbp]
\centering
\includegraphics[width=0.993\textwidth,height=0.84\textheight,keepaspectratio]{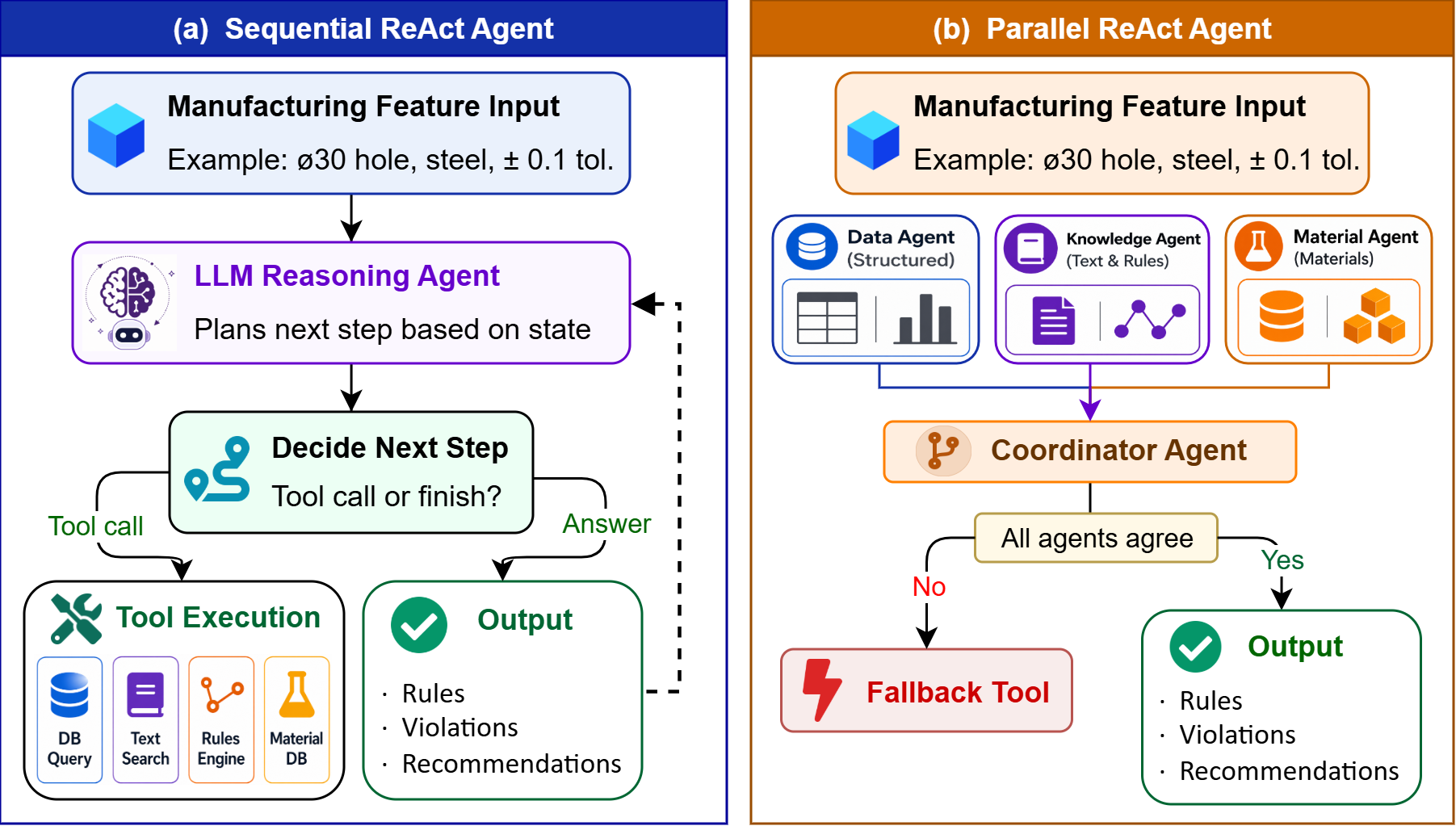}
\caption{Knowledge retrieval architecture, comparing sequential ReAct retrieval
using a single reasoning agent with parallel retrieval using three specialized
sub-agents and a coordination node.}
\end{figure}

\HeadingThree{3.4.1 Multi-Source Knowledge Base}

The KB integrates six complementary local knowledge modalities, each
representing a distinct form of manufacturing knowledge, as summarized in
Table 1. An external fallback mechanism is additionally provided for out-of-KB
cases where local sources are insufficient. This design reflects the
heterogeneity of manufacturing information: relational data supports efficient
numerical queries, text sources capture contextual reasoning, graph structures
encode relationships, and rule-based systems provide interpretable logic.

\begin{table}[!htbp]
\caption{Multi-source tool registry for knowledge retrieval.}
\centering
\small
\setstretch{1.15}
\begin{tabularx}{\textwidth}{|L{0.696}|L{0.612}|J{1.692}|}
\hline
\textbf{Tool} & \textbf{Modality} & \textbf{Content} \\
\hline
SQL retrieval & Relational database & Quantitative constraints, including
minimum diameters, aspect ratios, tolerance mappings, and thread requirements \\
\hline
Tabular retrieval & Structured data & Feature-material compatibility,
machinability scores, and coating recommendations \\
\hline
Text retrieval (RAG) & Text corpus & Engineering guidelines, process rationale,
and material practices \\
\hline
Decision tree evaluation & Rule-based logic & Feature classification and
constraint applicability \\
\hline
Knowledge graph query & Graph representation & Relationships between materials,
processes, and feature types \\
\hline
Material database lookup & Material properties & Machinability, thermal
behavior, hardness, and composition \\
\hline
External fallback & Generative model & Supplementary guidance for out-of-KB
queries after local sources are exhausted \\
\hline
\end{tabularx}
\end{table}

A shared normalization component handles variability in input terminology by
mapping informal, misspelled, or non-standard inputs to canonical forms. The
normalization follows a staged process: exact matching, partial matching,
model-based mapping, and fallback matching. Results are cached in memory to
improve efficiency for repeated queries. This mechanism is essential for
real-world inputs, where variations such as ``al-6061-t6'' and ``Aluminum 6061''
must be resolved to a consistent identifier prior to retrieval.

\HeadingThree{3.4.2 Sequential and Parallel ReAct Architectures}

The sequential configuration implements a ReAct loop within a stateful execution
graph, as shown in Fig. 6(a). At each step, the reasoning agent receives the
current feature context, previously retrieved evidence, and available tool
descriptions. It then decides whether to invoke another knowledge tool or
generate a final structured response. Retrieved tool results are appended to the
agent state, allowing subsequent reasoning steps to account for previously
accessed evidence. This process continues until sufficient manufacturing
knowledge has been collected or a stopping condition is reached. The sequential
design provides a complete reasoning trajectory and is useful for cases requiring
deeper validation, conflict checking, or step-by-step refinement.

A key component of the sequential KR Agent is its system prompt, which
constrains LLM reasoning within a manufacturing-specific tool-use workflow
rather than allowing unrestricted generation. As shown in Fig. 7, the prompt
encodes four functional elements. First, it defines the ReAct reasoning loop,
requiring the agent to alternate between reasoning, tool selection, tool
observation, and final response generation. Second, it provides
feature-type-specific tool-selection guidance so that dimensional limits,
tolerance requirements, material constraints, and process recommendations are
routed to appropriate knowledge sources. Third, it defines a source-priority
hierarchy for resolving conflicting information across structured databases,
tabular rules, textual guidelines, knowledge graphs, and material databases.
Fourth, it includes an external-knowledge guard that prevents premature fallback
to broader generative knowledge before relevant local sources have been queried.
This prompt design is central to the KR Agent because it enables the LLM to
operate as an interactive reasoning agent that retrieves, compares, and
synthesizes manufacturing evidence instead of generating unsupported
recommendations.

\begin{figure}[!htbp]
\centering
\includegraphics[width=\textwidth,height=0.84\textheight,keepaspectratio]{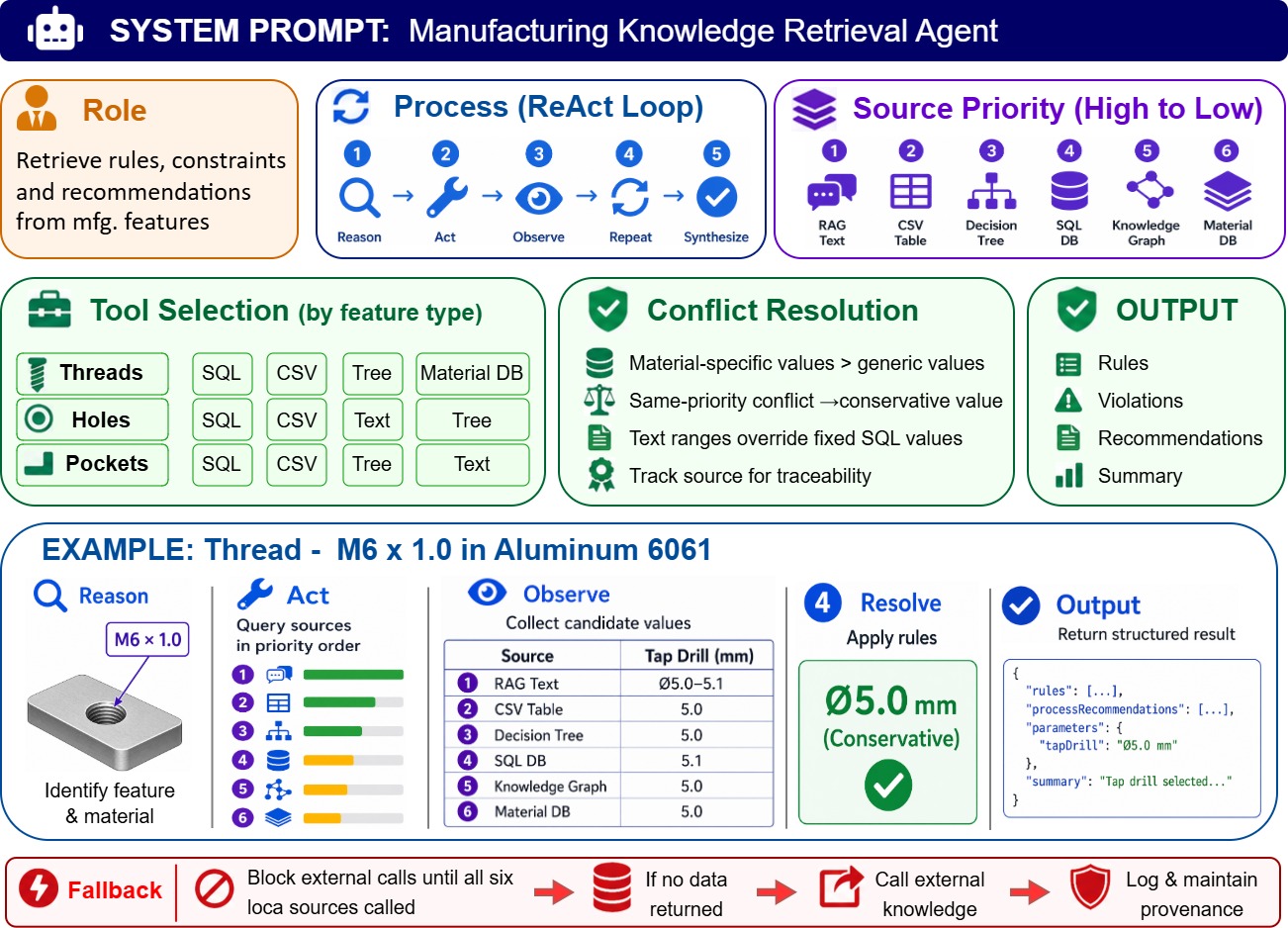}
\caption{Sequential ReAct prompt design for the Knowledge Retrieval Agent,
defining the reasoning loop, feature-specific tool-use guidance, source-priority
rules, and fallback constraints for grounded retrieval.}
\end{figure}

The parallel configuration distributes the same tool-grounded reasoning
objective across three specialized sub-agents, as shown in Fig. 6(b). The
structured-data sub-agent handles relational, tabular, and rule-based queries;
the text-knowledge sub-agent retrieves information from textual guidelines and
knowledge-graph sources; and the material sub-agent retrieves material-specific
properties and constraints. Each sub-agent follows a restricted ReAct loop
within its assigned tool scope, reducing unnecessary context accumulation while
improving coverage of heterogeneous knowledge sources. A coordination node then
merges the sub-agent outputs, applies the same source-priority logic used in the
sequential configuration, resolves conflicts where possible, and generates a
unified response with provenance and confidence information. The condensed
prompts for the parallel sub-agents and coordinator are provided in
\textbf{Appendix}.

This design allows the KR Agent to support both deep sequential reasoning and
broad parallel evidence collection. The sequential architecture emphasizes
reasoning continuity and iterative refinement, while the parallel architecture
emphasizes source coverage, fault tolerance, and efficiency. In both
configurations, the prompt-guided ReAct workflow ensures that manufacturing
recommendations remain grounded in retrieved evidence, traceable to specific
knowledge sources, and constrained by domain-specific decision rules.

\HeadingTwo{3.5 Process Sequence Agent}

The Process Sequence (PS) Agent receives the unified feature context from the
Context Fusion Agent and manufacturing constraints from the KR Agent. It
generates an ordered sequence of manufacturing processes required to produce the
part, acting as the bridge between knowledge interpretation and production
planning. Based on extracted constraints and feature information, the agent
determines the required operations, their ordering, and the associated
rationale. The task is inherently combinatorial: for parts with multiple
features, each requiring one or more operations, the number of valid sequences
is constrained by dependencies, material compatibility, and feature
accessibility. Rather than performing exhaustive enumeration, the agent applies
a deterministic decision hierarchy that prioritizes authoritative specifications
and generates sequences only when explicit definitions are unavailable.

\HeadingThree{3.5.1 Decision Logic}

The agent follows a three-path decision strategy with strict priority ordering,
as shown in Fig. 8.

\begin{wordbullets}
\item \textbf{Path 1 (User override):} If a HITL correction specifies a process
sequence, it is validated and returned with highest priority, as it reflects
expert input incorporating contextual factors not captured by the system.

\item \textbf{Path 2 (Design-specified process):} If CAD or drawing inputs
include process annotations, these are validated against known constraints and
returned as the primary output.

\item \textbf{Path 3 (Generated sequence):} If no explicit specification exists,
the agent generates a sequence using three knowledge structures. Process
templates define standard workflows for common part categories, such as machined
components, shafts, sheet metal, and cast parts. Material-process compatibility
mappings restrict feasible processes based on material properties, while process
dependency rules enforce valid precedence relationships between operations.
Sequences generated through this path represent the system's recommended
manufacturing plan.
\end{wordbullets}

\begin{figure}[!htbp]
\centering
\includegraphics[width=0.84\textwidth,height=0.84\textheight,keepaspectratio]{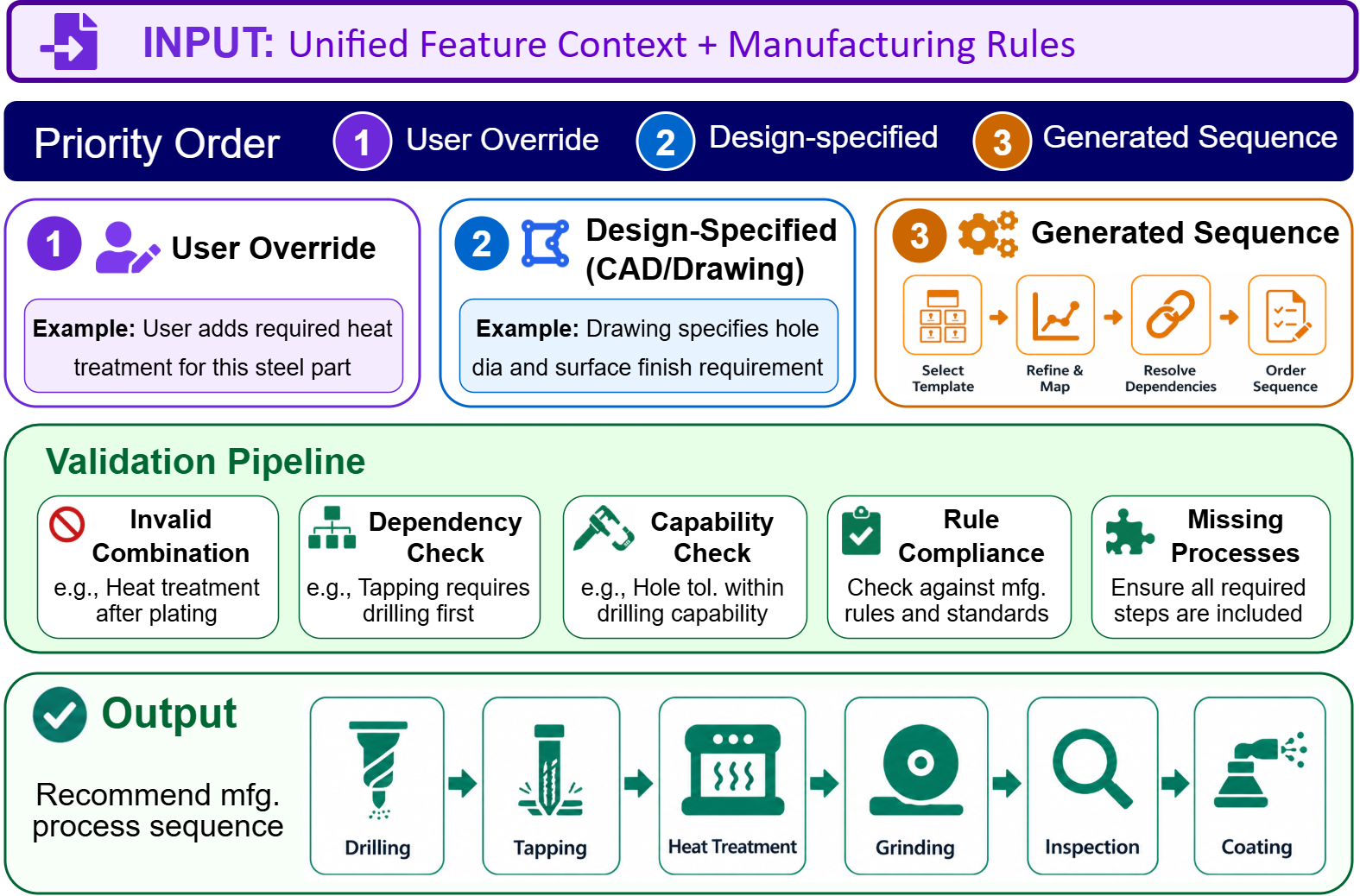}
\caption{Process sequence decision and validation workflow, showing
priority-based selection of user-defined, design-specified, or generated
sequences using templates, material compatibility, dependency rules, and
validation checks.}
\end{figure}

\HeadingThree{3.5.2 Validation and Confidence Scoring}

All generated or provided sequences are validated against multiple constraint
categories, including invalid process combinations, ordering violations, process
capability mismatches, and compliance with manufacturing rules. Validation
results are used to compute a confidence score:

\begin{equation}
C_{seq} = C_{base} - 0.3 \cdot I_{violation} - 0.1 \cdot \min(n_{warn}/5,\ 1.0)
\end{equation}

where $C_{base}$ is the base confidence associated with the sequence source,
$I_{violation}$ indicates the presence of a critical violation, and $n_{warn}$
is the number of warnings. The base confidence reflects the source hierarchy,
with user-defined and design-specified sequences assigned higher initial
confidence than generated ones. The penalty terms are used as an interpretable
validation heuristic rather than a probabilistic uncertainty model: critical
violations receive a larger penalty because they can invalidate a sequence,
while warnings reduce confidence gradually up to a capped limit. The score is
bounded within a fixed range to ensure interpretability: higher values indicate
valid and authoritative sequences, while lower values indicate inconsistencies
requiring review.

When validation identifies issues, the agent generates alternative sequences for
comparison. For complex or non-standard cases, a ReAct-based extension performs
iterative reasoning using specialized tools for process capability lookup,
sequence retrieval, rule evaluation, and knowledge search. The PS and TS agents
share this reasoning framework, differing only in domain-specific tools and
prompts.

\HeadingTwo{3.6 Tool Selection Agent}

The Tool Selection (TS) Agent receives the validated process sequence and
selects appropriate cutting tools and machining parameters for each operation,
as shown in Fig. 9. It translates process-level decisions into operation-level
manufacturing instructions, bridging process planning and machining execution.

\HeadingThree{3.6.1 Selection Logic and Tool Library}

The agent maintains a structured tool library organized into categories such as
endmills, drills, taps, boring tools, reamers, thread mills, grinding tools,
chamfer tools, and specialized tooling. Each entry includes material
compatibility, applicable processes, geometric constraints, coating type, and
tolerance capability. For each process step, a four-stage selection procedure is
applied. First, a tooling requirement check determines whether a cutting tool is
needed. Second, category identification retrieves relevant tool classes. Third,
candidate filtering removes incompatible tools based on material, geometry, and
tolerance constraints. Finally, a primary tool is selected based on feature
characteristics, surface requirements, and production considerations.
Process-to-tool mappings link each operation to valid tool categories, ensuring
only applicable tools are considered.

\HeadingThree{3.6.2 Parameter Calculation}

For each selected tool, machining parameters are computed using
material-dependent reference values. Spindle speed is given by:

\begin{equation}
N = (V_{c} \times 1000)/(\pi \times D)
\end{equation}

where $N$ is spindle speed, $V_{c}$ is cutting speed, and $D$ is tool diameter.

The feed rate for milling operations is:

\begin{equation}
f_{m} = f_{z} \times z \times N
\end{equation}

where $f_{m}$ is table feed rate, $f_{z}$ is feed per tooth, $z$ is the number
of cutting edges, and $N$ is spindle speed. A representative cutting speed is
selected within recommended ranges for the material-tool combination, with
adjustment factors applied based on tool characteristics such as coating. Depth
of cut is determined by operation type, with larger values for roughing and
smaller values for finishing. The width of cut is defined as a fraction of tool
diameter, with higher values for roughing and lower values for finishing.
Specialized operations follow additional constraints; for example, tapping feed
is synchronized with thread pitch, and reaming uses conservative parameters to
ensure dimensional accuracy.

\begin{figure}[!htbp]
\centering
\includegraphics[width=0.84\textwidth,height=0.84\textheight,keepaspectratio]{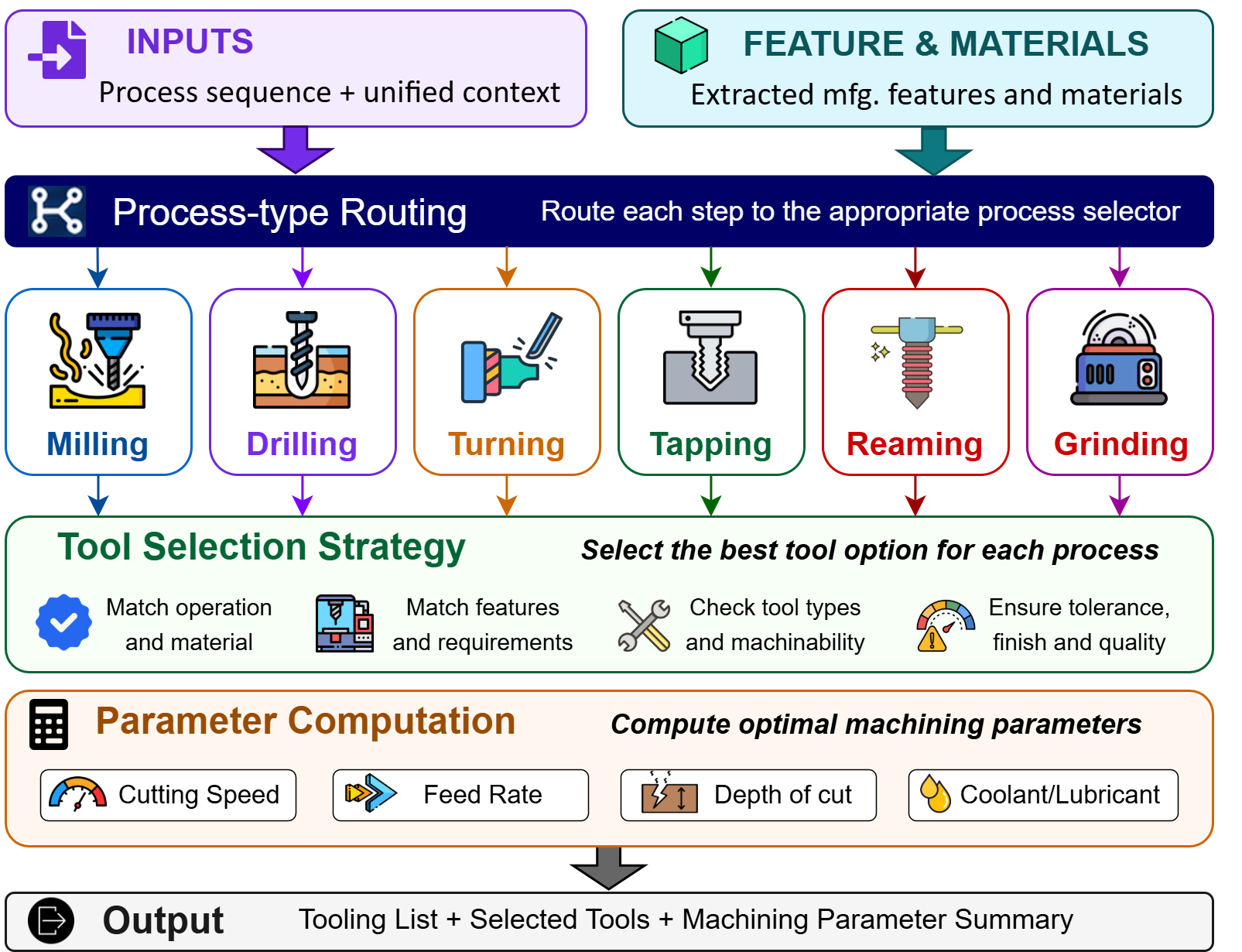}
\caption{Tool selection and parameter computation workflow, assigning tools
based on process, material, geometry, and tolerance constraints, with machining
parameters computed from material-dependent reference values.}
\end{figure}

\HeadingThree{3.6.3 Confidence Scoring}

The tool selection confidence score is designed as a coverage-based heuristic to
quantify whether all tooling-required process steps have been assigned valid
tools. It is not intended to represent probabilistic uncertainty; rather, it
provides an interpretable completeness indicator for downstream reporting and
human review:

\begin{equation}
C_{tool} = 0.3 + 0.7 \times (n_{tooled}/n_{required})
\end{equation}

where $n_{tooled}$ is the number of process steps with assigned tools and
$n_{required}$ is the total number of steps requiring tooling. The constant 0.3
is used as a minimum baseline to indicate that a process-level tooling
requirement has been identified even when tool assignment is incomplete, while
the remaining 0.7 is allocated to assignment coverage. This weighting ensures
that the score increases monotonically with tool assignment completeness and
reaches 1.0 only when all required tooling steps are specified. Lower scores
indicate partial tool coverage and therefore trigger manual review or further
reasoning. For non-standard materials, complex geometries, or unsupported
processes, a ReAct-based extension performs iterative reasoning using
specialized tools for parameter estimation, tool selection, and constraint
evaluation. The TS Agent shares the same reasoning framework as the PS and KR
agents, differing only in domain-specific tools and knowledge sources.

\HeadingTwo{3.7 Report Generation Agent}

The Report Generation Agent synthesizes outputs from all upstream agents into
structured manufacturing analysis reports. It uses template-based narration
instead of open-ended generation, ensuring consistency, reproducibility, and
auditability. This design prioritizes deterministic and verifiable
documentation, which is essential in engineering applications. Reports are
organized into seven sections: Executive Summary, Design Specifications,
Manufacturing Rules, Process Plan, Tool Selection, Risk Analysis, and
Recommendations. Multiple levels of detail, including executive, intermediate,
standard, and detailed views, and export formats, including web-viewable,
print-ready, editable, and structured data formats, are supported, enabling
integration with both human workflows and downstream computational systems.

\HeadingOne{4. Evaluation Methodology}

\HeadingTwo{4.1 Framework Design}

The evaluation framework assesses agent behavior across the range of inputs
expected in realistic manufacturing scenarios, including both standard and
complex conditions. A ground-truth benchmark is constructed with explicit
difficulty stratification and category labeling, where each test case is
assigned to a predefined complexity category. This design enables systematic
analysis of performance across varying levels of input difficulty.

For the three knowledge-intensive agents, namely KR, PS, and TS, evaluation is
conducted using a dedicated evaluation module under controlled input conditions.
This setup isolates agent-level behavior from workflow-level variability and
enables direct comparison of sequential and parallel configurations using the
same benchmark cases. For each test case, three artifacts are recorded: (i) a
complete execution trace, including tool calls, input arguments, retrieved
results, intermediate reasoning steps, and token usage; (ii) quantitative
evaluation metrics; and (iii) a human-readable summary. This structured logging
supports post-hoc analysis, debugging, and reproducibility of experimental
results.

\HeadingTwo{4.2 Knowledge Retrieval Benchmark}

The benchmark consists of 110 test cases: a primary dataset of 100 cases
spanning ten input-complexity categories, and an additional set of 10 cases
designed to evaluate conflict detection. The distribution of the primary dataset
reflects the expected frequency and difficulty of inputs in practical
manufacturing scenarios, as shown in Table 2. Each category represents a
distinct class of real-world challenges, enabling systematic evaluation of
robustness.

\begin{table}[!htbp]
\caption{Ground-truth dataset for knowledge retrieval: category distribution of
the 100 primary cases.}
\centering
\small
\setstretch{1.15}
\begin{tabularx}{\textwidth}{|L{0.724}|L{0.336}|J{2.484}|L{0.456}|}
\hline
\textbf{Category} & \textbf{Count} & \textbf{Description} & \textbf{Difficulty} \\
\hline
Normal & 18 & Complete, well-formed inputs with standard materials and features & Easy \\
\hline
Violation & 14 & Inputs violating manufacturing constraints & Medium \\
\hline
Edge Case & 8 & Boundary values and non-standard parameters & Hard \\
\hline
Missing Data & 5 & Materials or features absent from local database & Hard \\
\hline
Minimal Input & 10 & Feature type only, without parameters & Hard \\
\hline
Partial Input & 8 & Feature type with incomplete parameters & Hard \\
\hline
Ambiguous Input & 7 & Vague or informal feature naming & Hard \\
\hline
Messy Format & 15 & Misspellings, trade names, formatting inconsistencies & Hard \\
\hline
Outside KB & 10 & Queries with no relevant local data & Hard \\
\hline
Cross-Domain & 5 & Queries spanning multiple features or materials & Hard \\
\hline
\textbf{Total} & \textbf{100} &  &  \\
\hline
\end{tabularx}
\end{table}

Each test case defines input parameters, expected tool usage, including
mandatory and optional tools, and expected output properties. Output properties
include minimum rule coverage, severity levels, detected violations, and the
presence of process recommendations. An additional set of 10 cases evaluates
conflict detection behavior. These cases incorporate 13 known inconsistencies
across the six local knowledge sources. The conflicts fall into two categories:
(i) machinability discrepancies across materials, including aluminum, steel,
stainless steel, titanium, and plastics, where values differ between tabular and
material databases; and (ii) design constraint inconsistencies, including thread
engagement ratios, minimum diameters, fillet radii, and countersink angles,
where values differ between structured databases and updated standards. This
extended dataset enables explicit evaluation of the agent's ability to detect,
reason about, and resolve conflicting information across heterogeneous sources.

\HeadingTwo{4.3 Evaluation Metrics}

The primary metrics for evaluating tool usage are precision, recall, and F1
score. Let $T_{called}$ denote the set of tools invoked by the agent,
$T_{expected}$ the required tools, and $T_{optional}$ the acceptable optional
tools. The relevant tool set for precision is defined as:

\begin{equation}
T_{relevant} = T_{expected} \cup T_{optional}
\end{equation}

The metrics are computed as:

\begin{align}
P_{tool} &= |T_{called} \cap T_{relevant}|/|T_{called}| \\
R_{tool} &= |T_{called} \cap T_{expected}|/|T_{expected}| \\
F1_{tool} &= (2 \times P_{tool} \times R_{tool})/(P_{tool} + R_{tool})
\end{align}

Precision considers both required and optional tools, whereas recall considers
only required tools. This asymmetric formulation reflects manufacturing
semantics: multiple tools may provide equivalent information and should not
penalize precision, whereas missing required tools results in incomplete
reasoning and must penalize recall.

Additional correctness metrics include:

\begin{wordbullets}
\item \textbf{Rules Sufficient Rate:} fraction of cases where retrieved rules
meet or exceed the minimum required threshold.

\item \textbf{Severity Accuracy:} fraction of cases where the detected severity
matches or exceeds the expected severity level.
\end{wordbullets}

Efficiency is measured using four metrics per case: iteration count, number of
tool calls, token usage, and execution time. These metrics capture both
computational cost and reasoning efficiency. The fallback mechanism is evaluated
using trigger rate, precision, and recall. Trigger rate denotes the fraction of
cases where fallback is activated, precision denotes the fraction of triggered
cases where fallback was expected, and recall denotes the fraction of expected
fallback cases correctly identified.

Conflict detection performance is measured using the Conflict Detection Score
(CDS), a rubric-based composite metric. The rubric consists of three components:
detection, resolution, and explanation. Detection evaluates whether relevant
conflicting sources are queried, resolution evaluates whether the correct
authoritative value is selected, and explanation evaluates whether the conflict
is explicitly acknowledged and contextualized. The composite score is defined
as:

\begin{equation}
CDS = 0.3 \cdot D + 0.4 \cdot R + 0.3 \cdot E
\end{equation}

where $D$, $R$, and $E$ denote detection, resolution, and explanation scores,
respectively. Detection measures the fraction of conflicting sources accessed,
resolution measures whether the correct authoritative value is selected, and
explanation measures whether the conflict is explicitly identified and
contextualized. The weighting reflects the relative importance of the three
components in manufacturing decision-making. Detection and explanation are
necessary for traceability, but resolution is assigned a slightly higher weight
because selecting an appropriate constraint value has the most direct effect on
downstream process planning and tool selection. The weights are therefore used
as an interpretable evaluation rubric rather than a statistically learned
parameter set. All CDS evaluations are performed by a single domain expert;
future work should include sensitivity analysis of the weighting scheme and
multiple expert annotators to further validate the metric.

\HeadingTwo{4.4 Process Sequence and Tool Selection Benchmarks}

The PS and TS agents are evaluated using dedicated 100-case benchmarks with
ten-category stratification following the same design used for the KR benchmark.
For TS, two categories are modified: out-of-KB and cross-domain are replaced by
special tooling, covering non-standard tools such as gun drills and form
cutters, and multi-process scenarios, covering coordinated tool selection across
multiple stages. The remaining categories are shared across agents, enabling
comparative analysis.

The PS and TS agents use GPT-4o-mini, while KR uses GPT-4o. This configuration
reflects a trade-off between computational cost and reasoning capability. KR
requires broader reasoning across heterogeneous sources, whereas PS and TS
operate in more constrained decision spaces supported by structured tools, such
as lookup tables and dependency graphs. The model backbones are reported to
support reproducibility and to clarify the interpretation of cross-agent
efficiency results. Efficiency comparisons across agents should therefore be
interpreted cautiously, as differences may reflect model capability in addition
to task complexity.

\textbf{Process Sequence Metrics}

Let $P_{gen}$ denote generated processes and $P_{exp}$ denote expected
processes. Process set overlap is measured using Jaccard similarity:

\begin{equation}
J_{proc} = |P_{gen} \cap P_{exp}| \, / \, |P_{gen} \cup P_{exp}|
\end{equation}

Process Completeness (PC) and Process Precision (PP) are defined as:

\begin{align}
PC &= |P_{gen} \cap P_{exp}| \, / \, |P_{exp}| \\
PP &= |P_{gen} \cap P_{exp}| \, / \, |P_{gen}|
\end{align}

PC measures coverage of expected processes, while PP measures correctness of
generated processes. Ordering quality is evaluated using Kendall's tau ($\tau$)
over shared processes, ranging from $-$1 for reversed order to +1 for perfect
agreement. Dependency Compliance measures the fraction of process pairs
satisfying precedence constraints. The Alternative Generation Rate captures the
proportion of cases where at least one alternative sequence is produced.

\textbf{Tool Selection Metrics}

Let $T_{gen}$ and $T_{exp}$ denote generated and expected tool sets. Tool Type
Jaccard follows the same formulation as above. Additional metrics include:

\begin{wordbullets}
\item \textbf{Material Compatibility Score:} fraction of tools satisfying
material constraints.

\item \textbf{Coating Score:} alignment between selected coatings and expected
specifications.

\item \textbf{Parameter Accuracy:} fraction of machining parameters within
acceptable tolerance ranges.

\item \textbf{Special Tooling Recall:} fraction of required non-standard tools
correctly identified.

\item \textbf{Must-Include Coverage:} fraction of mandatory tool specifications
present.
\end{wordbullets}

\textbf{System-Level Evaluation}

System-level evaluation focuses on integration of feature extraction and context
fusion. The GCNN service is evaluated using labeled STEP datasets for feature
recognition accuracy. Context fusion is evaluated by the proportion of 3D
features successfully mapped to corresponding 2D annotations based on
expert-validated ground truth. This metric reflects the effectiveness of
semantic interpretation and hybrid matching.

All ReAct evaluations are conducted as single-run experiments with temperature
set to zero, yielding near-deterministic outputs. Deterministic components
operate with fixed parameters. Reported metrics are point estimates from a
single evaluation pass. Some categories contain limited samples, with 5--7
cases, where single-case variations may significantly affect results.

\HeadingOne{5. Results and Discussion}

\HeadingTwo{5.1 Feature Extraction and Context Fusion}

Tables 3--5 summarize the performance of the upstream components that provide
structured inputs to downstream agents. Detailed evaluation procedures are
reported in \cite{khan2024automaticfeature} and \cite{khan2025multistagehybrid}; key results are summarized here.

\begin{table}[!htbp]
\caption{GCNN-based CAD feature recognition performance.}
\centering
\small
\setstretch{1.15}
\begin{tabular}{|p{5.0cm}|p{6.0cm}|}
\hline
\textbf{Metric} & \textbf{Value} \\
\hline
Overall Accuracy & 96.87\% \\
\hline
Precision & 97.16\% \\
\hline
Recall & 96.67\% \\
\hline
F1 Score & 96.87\% \\
\hline
Dimension Extraction Accuracy & 100\% (for correctly identified features) \\
\hline
Feature Classes & 36 (subtractive and additive) \\
\hline
Training Dataset & 150,000 synthetic CAD models \\
\hline
Inference Time & 3--4 s per model \\
\hline
\end{tabular}
\end{table}

The 36-class feature set covers common machining features for both prismatic and
rotational parts, including holes, pockets, slots, and steps. The B-Rep-based
graph representation effectively encodes geometric relationships, enabling the
hierarchical GCNN to achieve high classification accuracy with low inference
time. Accurate dimensional extraction for correctly identified features provides
a reliable foundation for downstream reasoning.

\begin{table}[!htbp]
\caption{Three-stage drawing analysis pipeline performance.}
\centering
\small
\setstretch{1.15}
\begin{tabular}{|>{\centering\arraybackslash}p{3.6cm}
                |>{\centering\arraybackslash}p{2.4cm}
                |>{\centering\arraybackslash}p{3.6cm}
                |>{\centering\arraybackslash}p{2.4cm}|}
\hline
\textbf{Stage} & \textbf{Model} & \textbf{Target} & \textbf{Accuracy / F1} \\
\hline
\multirow{3}{*}{Layout Detection} & \multirow{3}{*}{YOLOv11-det} & Views & 0.96 \\
\cline{3-4}
 &  & Title Block & 0.99 \\
\cline{3-4}
 &  & Notes & 0.98 \\
\hline
\multirow{3}{*}{Annotation Localization} & \multirow{3}{*}{YOLOv11-obb} & Measures & 0.95 \\
\cline{3-4}
 &  & GD\&T & 0.97 \\
\cline{3-4}
 &  & Surface Roughness & 0.54 \\
\hline
\multirow{4}{*}{Numerical Extraction} & \multirow{4}{*}{Donut VLM} & Measures (F1) & 0.923 \\
\cline{3-4}
 &  & GD\&T (F1) & 0.965 \\
\cline{3-4}
 &  & Surface Roughness (F1) & 1.000 \\
\cline{3-4}
 &  & Overall (F1) & 0.963 \\
\hline
\end{tabular}
\end{table}

Lower localization performance for surface roughness annotations is mainly
associated with dataset imbalance, highlighting sensitivity to class
distribution. The pipeline employs a hybrid strategy: numerical annotations are
processed using a document understanding transformer, while textual regions are
handled by a VLM. This separation improves robustness for both structured data
and free-form text. Replacing a generic document model with a VLM significantly
improved extraction of materials, notes, and metadata across diverse formats.
End-to-end processing requires approximately 2--5 s per drawing page; combined
with CAD analysis, total extraction time remains within 6--10 s due to parallel
execution.

\begin{table}[!htbp]
\caption{Context fusion mapping performance using 20 CAD--drawing pairs and 101
mappings.}
\centering
\small
\setstretch{1.15}
\begin{tabularx}{\textwidth}{|L{2.095}|L{0.730}|L{0.690}|L{0.690}|L{0.795}|}
\hline
\textbf{Metric} & \textbf{Mean} & \textbf{Std} & \textbf{Min} & \textbf{Max} \\
\hline
Mapping Precision & 0.837 & 0.170 & 0.50 & 1.00 \\
\hline
Mapping Recall & 0.905 & 0.117 & 0.57 & 1.00 \\
\hline
Mapping F1 Score & 0.863 & 0.137 & 0.57 & 1.00 \\
\hline
Exact Match Rate & 0.792 & 0.200 & 0.40 & 1.00 \\
\hline
Partial Match Rate & 0.903 & 0.123 & 0.57 & 1.00 \\
\hline
Inference Time (s) & 54.94 & 20.65 & 30.05 & 104.63 \\
\hline
\end{tabularx}
\end{table}

The context fusion component is evaluated on 20 real CAD--drawing pairs
comprising 101 feature-to-annotation mappings with expert-validated ground
truth. Approximately 40\% of features involve repeated or patterned instances,
increasing ambiguity. The deterministic-first pipeline resolves a subset of
mappings directly through scoring, while most cases require hybrid processing,
where candidate sets are generated deterministically and refined through
reasoning-based disambiguation. The high partial match rate indicates stable
behavior under ambiguity, as the system preserves multiple plausible mappings
rather than committing incorrect assignments. Ablation results for the context
fusion module show that removing domain-specific heuristics reduces precision,
while removing reasoning-based disambiguation reduces recall. The full pipeline
achieves the highest F1 score, demonstrating the effectiveness of the hybrid
deterministic-agentic design for multi-modal integration.

\HeadingTwo{5.2 Knowledge Retrieval Results}

Fig. 10 summarizes the aggregate performance of the sequential and parallel
ReAct configurations across the 100-case knowledge retrieval benchmark.

\begin{figure}[!htbp]
\centering
\includegraphics[width=0.948\textwidth,height=0.84\textheight,keepaspectratio]{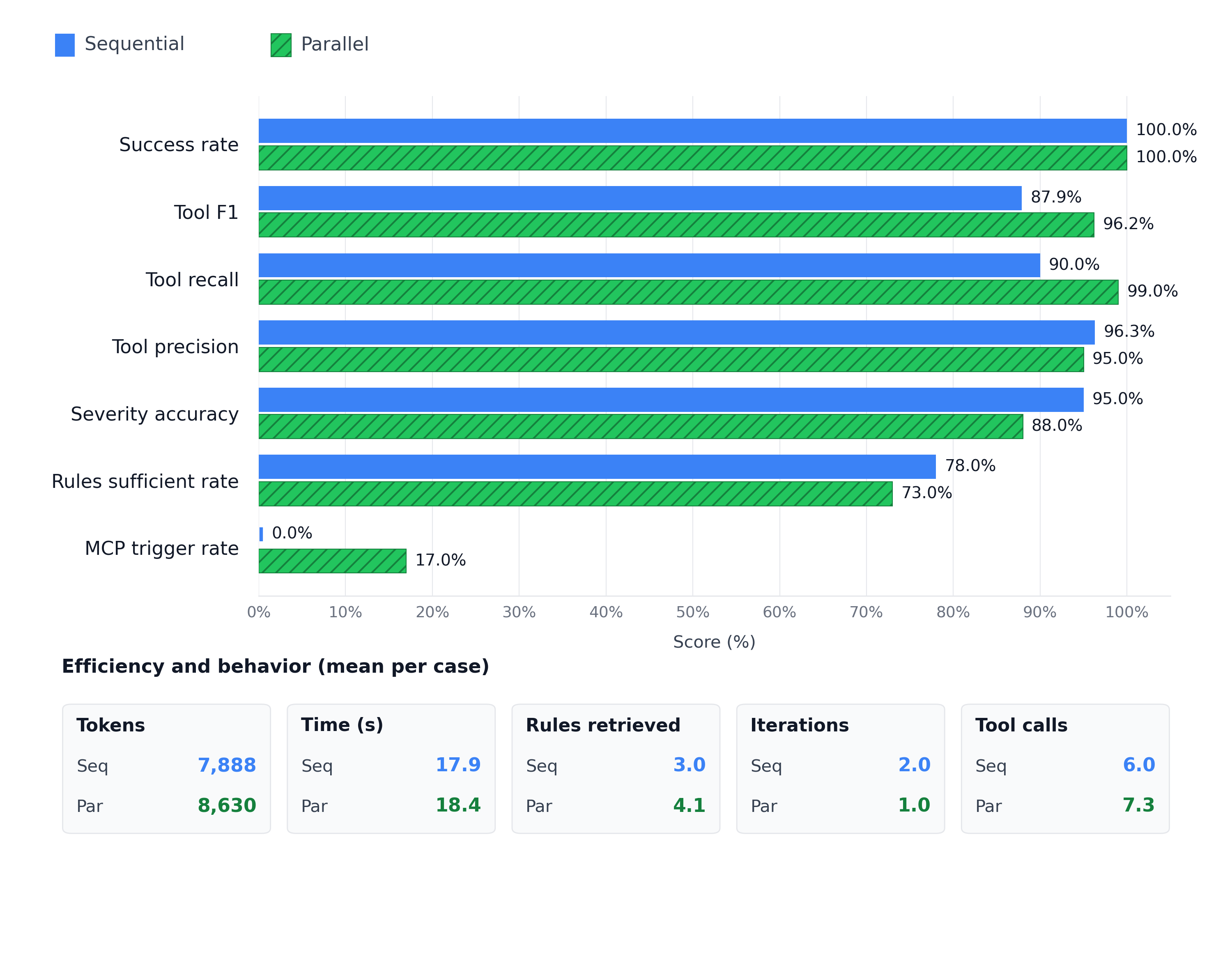}
\captionsetup{labelsep=space}
\caption{Performance comparison of sequential and parallel ReAct
configurations, showing higher Tool F1 for parallel retrieval and higher severity
accuracy for sequential reasoning.}
\end{figure}

Both configurations complete all benchmark cases, indicating that ReAct-based
retrieval with multi-source tools can operate reliably across noisy, ambiguous,
and out-of-KB inputs. The sequential configuration follows a deeper reasoning
pattern, averaging multiple tool-use steps for retrieval and refinement. In
contrast, the parallel configuration distributes retrieval across specialized
sub-agents, achieving broader source coverage with some redundancy.

The parallel configuration achieves higher Tool F1, primarily due to improved
recall, while the sequential configuration achieves slightly higher precision by
invoking fewer unnecessary tools. This reflects the expected architectural
trade-off: parallel reasoning improves coverage across heterogeneous sources,
whereas sequential reasoning maintains a more focused reasoning trajectory.
Severity accuracy is higher in the sequential configuration, suggesting that a
unified reasoning context can improve calibration of constraint severity,
particularly in edge-case and violation scenarios.

Fallback behavior further differentiates the two configurations. The sequential
configuration does not trigger fallback, indicating limited sensitivity to
incomplete source coverage. The parallel configuration activates fallback in
cases where local sources are insufficient, showing better detection of missing
information. However, this broader coverage introduces modest coordination
overhead, resulting in slightly higher token consumption for KR. Conflict
detection performance is summarized in Fig. 11.

\begin{figure}[!htbp]
\centering
\includegraphics[width=\textwidth,height=0.84\textheight,keepaspectratio]{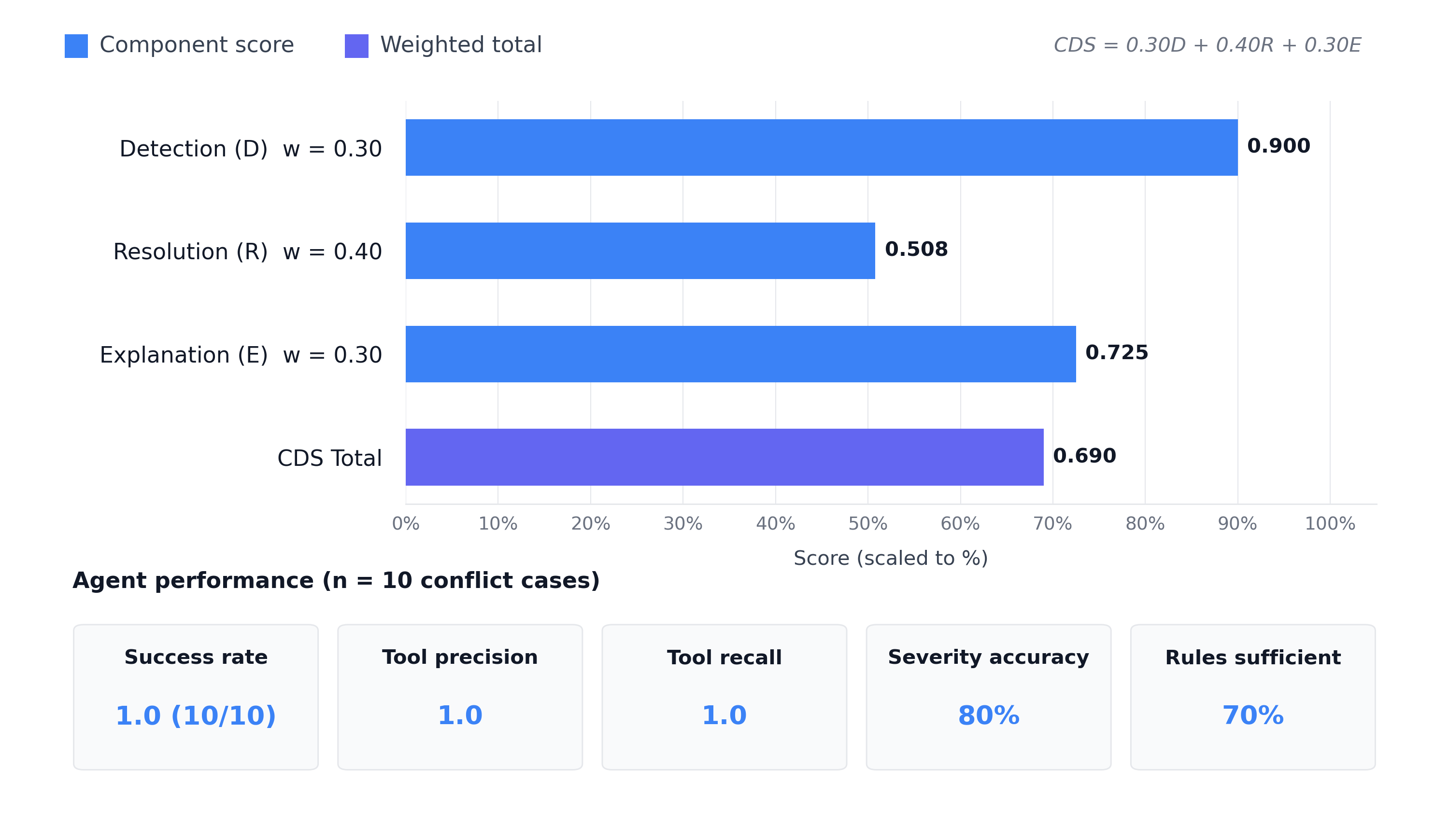}
\caption{Conflict Detection Score (CDS) breakdown, showing strong conflict
detection, moderate explanation quality, and weaker conflict resolution
performance.}
\end{figure}

The CDS results show that the agent is generally effective at accessing
conflicting sources and identifying the presence of inconsistencies. However,
resolution remains the weakest component, indicating that selecting a single
authoritative value is more difficult than detecting or explaining a conflict.
This behavior reflects a conservative reasoning strategy, where the agent may
present multiple plausible values rather than commit to one source when evidence
is inconsistent.

Per-case analysis in Table 6 further illustrates this behavior. Higher-scoring
cases occur when all relevant sources are queried and the selected value is
justified with an appropriate explanation. Moderate and lower-scoring cases are
mainly associated with incomplete source coverage, where one or more relevant
sources are not invoked.

\begin{table}[!htbp]
\caption{Representative CDS cases.}
\centering
\small
\setstretch{1.15}
\begin{tabularx}{\textwidth}{|L{0.606}|L{0.348}|L{0.438}|L{0.438}|L{0.438}|J{3.732}|}
\hline
\textbf{Case} & \textbf{D} & \textbf{R} & \textbf{E} & \textbf{CDS} & \textbf{Conflict Description} \\
\hline
CDS-001 & 1.00 & 0.50 & 0.75 & 0.725 & Thread engagement in aluminum: both
sources queried; conservative value selected \\
\hline
CDS-002 & 0.75 & 0.50 & 0.75 & 0.650 & Minimum hole diameter in steel: one
conflicting source missed \\
\hline
CDS-010 & 0.50 & 0.50 & 0.50 & 0.500 & Machinability of SS304: conflict missed
due to incomplete querying \\
\hline
\textbf{Mean} & \textbf{0.90} & \textbf{0.508} & \textbf{0.725} & \textbf{0.690} &  \\
\hline
\end{tabularx}
\end{table}

Overall, these results indicate that effective conflict detection depends
strongly on comprehensive tool invocation. The primary failure mode is
incomplete source access rather than incorrect reasoning after evidence has been
retrieved. This suggests that future improvements should emphasize prompt
constraints or retrieval policies that require cross-source comparison for
conflict-sensitive parameters, such as machinability values, engagement ratios,
and dimensional limits.

\HeadingTwo{5.3 Process Sequence and Tool Selection}

Both agents are evaluated on their respective 100-case benchmarks using
GPT-4o-mini. Aggregate results for the sequential and parallel configurations
are presented in Tables 7 and 8.

\begin{table}[!htbp]
\caption{Process Sequence Agent performance using 100 cases and GPT-4o-mini.}
\centering
\small
\setstretch{1.15}
\begin{tabularx}{\textwidth}{|L{1.794}|L{0.672}|L{0.531}|}
\hline
\textbf{Metric} & \textbf{Sequential} & \textbf{Parallel} \\
\hline
Success Rate & 99.0\% & 100.0\% \\
\hline
Tool F1 & 0.946 & 0.959 \\
\hline
Process Jaccard & 0.236 & 0.208 \\
\hline
Order (Kendall $\tau$) & 0.313 & 0.147 \\
\hline
Dependency Compliance & 0.863 & 0.888 \\
\hline
Process Completeness & 0.598 & 0.455 \\
\hline
Process Precision & 0.277 & 0.251 \\
\hline
Alternative Rate & 83.0\% & 100.0\% \\
\hline
Avg. Iterations & 5.4 & 1.0 \\
\hline
Avg. Tokens & 22,481 & 8,896 \\
\hline
External Fallback Trigger Rate & 0.0\% & 0.0\% \\
\hline
\end{tabularx}
\end{table}

The PS results show a quality-efficiency trade-off. The parallel configuration
achieves slightly higher Tool F1 and a perfect success rate, while the
sequential configuration produces stronger process-level agreement with the
reference sequences, particularly in ordering fidelity and completeness. This
difference reflects the benefit of iterative reasoning in the sequential
configuration, where multiple tool interactions allow repeated validation and
refinement.

The parallel configuration is substantially more token-efficient, while the
sequential configuration produces more comprehensive plans. Moderate values for
Process Jaccard and Kendall's $\tau$ are mainly due to systematic
over-generation of valid auxiliary operations, such as deburring, cleaning, and
inspection, that are not included in the minimal ground-truth sequences.
Therefore, these low overlap-based scores should not be interpreted as direct
manufacturing invalidity. Instead, they indicate a limitation of single-reference
evaluation, where expanded but plausible process plans may be penalized despite
satisfying manufacturing precedence constraints. This motivates future use of
multi-reference ground truth or expert-rated evaluation.

Neither configuration triggers external fallback, suggesting that the available
process knowledge tools provide sufficient coverage for the evaluated cases. The
parallel configuration also generates alternative sequences more consistently,
reflecting the diversity introduced by independent sub-agent proposals.

\begin{table}[!htbp]
\caption{Tool Selection Agent performance using 100 cases and GPT-4o-mini.}
\centering
\small
\setstretch{1.15}
\begin{tabularx}{\textwidth}{|L{1.794}|L{0.672}|L{0.531}|}
\hline
\textbf{Metric} & \textbf{Sequential} & \textbf{Parallel} \\
\hline
Success Rate & 96.0\% & 100.0\% \\
\hline
Tool F1 & 0.901 & 0.976 \\
\hline
Tool Type Jaccard & 0.639 & 0.678 \\
\hline
Material Compatibility & 0.726 & 0.702 \\
\hline
Coating Score & 0.350 & 0.457 \\
\hline
Parameter Accuracy & 0.910 & 0.940 \\
\hline
Special Tooling Recall & 0.860 & 0.870 \\
\hline
Must-Include Coverage & 0.665 & 0.723 \\
\hline
Avg. Iterations & 6.6 & 1.0 \\
\hline
Avg. Tokens & 36,659 & 11,868 \\
\hline
External Fallback Trigger Rate & 31.0\% & 0.0\% \\
\hline
\end{tabularx}
\end{table}

The TS results show a stronger advantage for the parallel configuration. It
achieves higher Tool F1, complete execution success, and better
coverage-oriented metrics, including Tool Type Jaccard, Parameter Accuracy, and
Must-Include Coverage. The sequential configuration shows slightly higher
material compatibility, suggesting that iterative validation can improve
material-specific checking, but this comes at a substantially higher token cost.

Coating selection remains the weakest aspect for both configurations, as
reflected by the relatively low coating scores. This indicates that coating
selection is underrepresented in the current tool library and requires richer
modeling of coating--substrate--material--operation interactions. These
decisions depend on material, tool substrate, cutting condition, and operation
type, which are not yet fully represented in the current tool library. In
contrast, parameter accuracy remains high, indicating that deterministic
parameter calculation produces reliable machining values once a suitable tool is
selected.

Fallback behavior differs from KR and PS. The sequential TS configuration
triggers fallback in a subset of difficult cases, indicating greater reliance on
external knowledge under uncertainty. The parallel configuration does not
trigger fallback, suggesting that distributed tool exploration provides
sufficient coverage within the available tool library. Overall, the TS results
show that parallel reasoning improves robustness and efficiency, while
sequential reasoning provides deeper validation at higher computational cost.

\HeadingTwo{5.4 Cross-Agent Analysis}

Table 9 summarizes key performance metrics across the three ReAct-enabled
agents. Sequential and parallel comparisons within each agent are directly
comparable because the backbone model is held constant for each agent. However,
cross-agent efficiency comparisons should be interpreted cautiously because KR
uses GPT-4o, whereas PS and TS use GPT-4o-mini.

\begin{table}[!htbp]
\caption{Cross-agent comparison of sequential and parallel architectures across
300 cases.}
\centering
\small
\setstretch{1.15}
\begin{tabularx}{\textwidth}{|L{2.226}|L{0.833}|L{0.833}|L{0.770}|L{0.770}|L{0.784}|L{0.784}|}
\hline
\textbf{Metric} & \textbf{KR Seq} & \textbf{KR Par} & \textbf{PS Seq} & \textbf{PS Par} & \textbf{TS Seq} & \textbf{TS Par} \\
\hline
Success Rate & 100\% & 100\% & 99\% & 100\% & 96\% & 100\% \\
\hline
Tool F1 & 0.879 & 0.962 & 0.946 & 0.959 & 0.901 & 0.976 \\
\hline
Avg. Iterations & 2.0 & 1.0 & 5.4 & 1.0 & 6.6 & 1.0 \\
\hline
Avg. Tokens & 7,888 & 8,630 & 22,481 & 8,896 & 36,659 & 11,868 \\
\hline
External Fallback Rate & 0\% & 17\% & 0\% & 0\% & 31\% & 0\% \\
\hline
\end{tabularx}
\end{table}

Several patterns emerge from the cross-agent results. The parallel architecture
achieves complete success across all agents and consistently improves Tool F1,
indicating stronger tool coverage and greater robustness from distributed
reasoning. The improvement is most pronounced in KR and TS, where broader source
or tool exploration is especially important.

Efficiency follows a task-dependent pattern. For KR, the parallel configuration
consumes slightly more tokens because coordination overhead offsets the benefit
of shorter reasoning paths. For PS and TS, however, the parallel configuration
substantially reduces token usage by replacing long sequential reasoning chains
with focused sub-agent execution. This suggests that the benefit of
parallelization increases as the sequential task requires more iterative
refinement.

Fallback behavior also differs by agent. In KR, fallback is activated only in
the parallel configuration, indicating better detection of insufficient local
knowledge coverage. In TS, fallback appears only in the sequential
configuration, suggesting that broader parallel tool exploration can reduce
reliance on external fallback. These results indicate that fallback behavior
depends on both task structure and the way evidence is distributed across
available tools.

\begin{table}[!htbp]
\caption{Per-category Tool F1 across agents.}
\centering
\small
\setstretch{1.15}
\begin{tabularx}{\textwidth}{|L{2.000}|L{0.528}|L{0.960}|L{0.952}|L{0.880}|L{0.880}|L{0.896}|L{0.896}|}
\hline
\textbf{Category} & \textbf{N} & \textbf{KR Seq} & \textbf{KR Par} & \textbf{PS Seq} & \textbf{PS Par} & \textbf{TS Seq} & \textbf{TS Par} \\
\hline
Normal & 18 & 0.941 & 0.937 & 0.944 & 0.989 & 0.900 & 1.000 \\
\hline
Violation & 14 & 0.974 & 0.973 & 1.000 & 0.986 & 0.912 & 1.000 \\
\hline
Edge Case & 8 & 0.927 & 0.918 & 0.990 & 1.000 & 0.821 & 1.000 \\
\hline
Missing Data & 5 & 0.982 & 0.978 & 0.985 & 1.000 & 0.954 & 1.000 \\
\hline
Minimal Input & 10 & 1.000 & 1.000 & 0.780 & 0.829 & 0.905 & 0.947 \\
\hline
Partial Input & 8 & 1.000 & 1.000 & 1.000 & 1.000 & 1.000 & 0.975 \\
\hline
Ambiguous & 7 & 1.000 & 0.989 & 0.768 & 0.813 & 0.845 & 0.939 \\
\hline
Messy Format & 15 & 1.000 & 0.974 & 0.995 & 0.973 & 0.948 & 0.987 \\
\hline
Outside KB & 10 & 0.000 & 0.900 & 0.969 & 1.000 & --- & --- \\
\hline
Cross-Domain & 5 & 1.000 & 0.985 & 1.000 & 0.943 & --- & --- \\
\hline
Special Tooling & 10 & --- & --- & --- & --- & 0.780 & 0.909 \\
\hline
Multi-Process & 5 & --- & --- & --- & --- & 0.954 & 0.971 \\
\hline
\end{tabularx}
\end{table}

The per-category results show that input normalization is effective, as all
agents maintain strong performance on messy-format cases. The largest
architectural difference appears in the outside-KB category, where sequential KR
fails to trigger the necessary fallback while parallel KR maintains high
performance. Minimal and ambiguous inputs remain more challenging because they
provide limited context for process and tool reasoning. For TS, special tooling
remains the most difficult category, reflecting limited coverage of non-standard
tools in the current library. Overall, the category-level results confirm that
parallel reasoning improves robustness in cases requiring broader tool coverage,
while remaining limitations are mainly associated with sparse input context and
incomplete tool-library coverage.

\HeadingTwo{5.5 Discussion}

The results show that the hybrid deterministic-agentic architecture combines
deterministic reliability with flexible LLM-based reasoning for manufacturing
process planning. Fully LLM-based perception would be slower, computationally
expensive, and less reliable due to possible hallucination of geometric
properties. Conversely, purely rule-based planning would struggle with the
majority of benchmark cases classified as difficult, which require reasoning
over ambiguous, incomplete, or conflicting inputs.

The system adopts a task-dependent autonomy profile. Feature extraction is
handled by deterministic perception modules, while downstream reasoning tasks
are assigned to LLM-based agents operating within constrained tool environments.
This design reflects the central objective of the framework: LLMs are not used
as standalone text generators, but as interactive reasoning agents that retrieve
information, compare heterogeneous sources, resolve uncertainty, and synthesize
manufacturing decisions. Across all agents, Tool F1 remains above 0.87,
indicating that bounded tool use effectively constrains LLM reasoning while
preserving flexibility.

Evaluation results provide implicit ablation signals that highlight the
contribution of individual components, as shown in Table 11. The normalization
module acts as a critical enabler for cases containing non-standard inputs,
ensuring consistent mapping to canonical database entries. The fallback
mechanism demonstrates the importance of external knowledge access: sequential
KR without fallback fails completely on out-of-KB cases, whereas parallel KR
with fallback achieves strong recovery. Similarly, the multi-source knowledge
architecture enables conflict awareness through source comparison and CDS-based
evaluation.

More broadly, the results illustrate the limitations of alternative approaches.
A single-LLM system would lack traceability to validated sources, numerical
precision from structured data, and reliable mechanisms for detecting
cross-source inconsistencies. A RAG-only system would be unable to access
structured knowledge sources such as relational databases and knowledge graphs.
Traditional rule-based systems would require manual intervention for many
difficult cases, whereas the proposed framework maintains high success rates
through coordinated agentic reasoning.

\begin{table}[!htbp]
\caption{Component contribution analysis.}
\centering
\small
\setstretch{1.15}
\begin{tabularx}{\textwidth}{|L{1.040}|L{0.875}|L{1.080}|L{1.005}|L{1.000}|}
\hline
\textbf{Component} & \textbf{Evidence Source} & \textbf{With Component} & \textbf{Without / Degraded} & \textbf{Impact} \\
\hline
LLM Normalizer & Messy Format cases & F1: 0.948-1.0 & Lookup failures for
non-standard inputs & Robustness to non-standard terminology \\
\hline
External Fallback & Out-of-KB cases & Parallel KR F1 = 0.90 & Sequential KR
F1 = 0.0 & Out-of-distribution coverage \\
\hline
Multi-Source Architecture & CDS cases & Detection = 0.90, CDS = 0.69 & No
conflict detection & Conflict awareness \\
\hline
VLM Semantic Interpretation & Context Fusion & Accurate disambiguation &
Numeric-only matching may cause false matches & Mapping accuracy \\
\hline
Parallel Consensus & All agents & 100\% success & 96-100\% success & Fault
tolerance \\
\hline
Tool-Grounded Reasoning & Structured tool-use cases & Tool F1: 0.879-0.976 & No
source traceability & Accuracy and auditability \\
\hline
\end{tabularx}
\end{table}

Table 12 positions the proposed system within the broader landscape of
multi-agent manufacturing systems. Prior work has demonstrated inter-agent
communication, particularly in holonic manufacturing architectures \cite{leitao2006adacor,kruger2019jadeerlang}, and
recent LLM-based systems have addressed selected manufacturing tasks. However,
these systems are typically limited to scheduling, shopfloor control, production
management, or other individual stages. In contrast, the proposed framework
provides end-to-end coverage of the process planning workflow by integrating
feature recognition, context fusion, knowledge retrieval, process sequencing,
tool selection, and report generation within a unified design-to-plan system.

This distinction is important because the main contribution of the proposed
framework is not only the use of multiple agents, but the use of agentic
coordination to bridge heterogeneous design representations and downstream
manufacturing decisions. The system connects 3D CAD-derived features with 2D
drawing-derived engineering specifications, enriches them into feature-level
manufacturing context, and uses this fused context to support tool-grounded
reasoning across knowledge retrieval, process planning, and tool selection. This
design directly addresses the core gap identified in the literature: the absence
of a fully implemented agentic framework that connects design information with
executable manufacturing planning outputs.

\begin{table}[!htbp]
\caption{Comparison of multi-agent manufacturing systems.}
\centering
\small
\setstretch{1.15}
\begin{tabular}{|>{\raggedright\arraybackslash}p{2.42cm}
                |>{\raggedright\arraybackslash}p{2.87cm}
                |>{\raggedright\arraybackslash}p{2.43cm}
                |>{\raggedright\arraybackslash}p{4.41cm}
                |>{\raggedright\arraybackslash}p{2.87cm}|}
\hline
\multicolumn{1}{|>{\centering\arraybackslash}p{2.42cm}|}{\textbf{System}} &
\multicolumn{1}{>{\centering\arraybackslash}p{2.87cm}|}{\textbf{Communication}} &
\multicolumn{1}{>{\centering\arraybackslash}p{2.43cm}|}{\textbf{Quantitative Evaluation}} &
\multicolumn{1}{>{\centering\arraybackslash}p{4.41cm}|}{\textbf{Pipeline Scope}} &
\multicolumn{1}{>{\centering\arraybackslash}p{2.87cm}|}{\textbf{Agents}} \\
\hline
PROSA \cite{vanbrussel1998prosa} & Holonic MAS / no fixed protocol & Conceptual or limited & Holonic
manufacturing control, including scheduling & 3 core holons + optional staff
holon \\
\hline
ADACOR \cite{leitao2006adacor} & Holonic MAS / JADE, FIPA-ACL & Yes, simulation experiments &
Adaptive shopfloor control & 4 holon types \\
\hline
MASCAPP \cite{nassehi2006stepnc} & MAS / agent messaging & Limited & CAPP for prismatic parts &
Multi-agent \\
\hline
Kruger \& Basson \cite{kruger2019jadeerlang} & MAS / JADE + Erlang & Yes & Holonic cell control & 3 \\
\hline
Xia et al. \cite{xia2023towardsautonomous} & LLM-augmented MAS + digital twin & Limited & Production
control & Hierarchical multi-tier agents \\
\hline
Zhao et al. \cite{zhao2026llmshopfloor} & MAS on physical system & Yes & Shopfloor scheduling &
5 modules \\
\hline
Liu et al. \cite{liu2026llmembodied} & Embodied agents & Yes & Scheduling and disturbance handling &
Per-machine agents \\
\hline
\textbf{This work} & \textbf{Structured asynchronous messaging} & \textbf{Yes} &
\textbf{End-to-end manufacturing process planning} & \textbf{6 agents + 2 ML
services} \\
\hline
\end{tabular}
\end{table}

The results also highlight complementary strengths of the sequential and
parallel ReAct architectures. The parallel configuration provides high
reliability, broad tool coverage, and strong efficiency, making it suitable for
deployment scenarios prioritizing robustness and cost efficiency. Substantial
token reductions for PS and TS directly translate into lower operational cost at
scale. The sequential configuration offers higher-quality reasoning in specific
dimensions, including severity calibration, process ordering fidelity, and
material compatibility. These advantages arise from iterative reasoning, which
enables deeper validation and refinement.

A hybrid deployment strategy is therefore recommended. The parallel
configuration can serve as the default execution mode, providing efficient and
robust baseline performance. The sequential configuration can be selectively
applied to cases requiring deeper reasoning, such as high-risk, ambiguous, or
safety-critical inputs. This strategy combines the throughput and fault
tolerance of parallel reasoning with the deeper validation capability of
sequential reasoning.

\HeadingTwo{5.6 Case Study}

To complement the quantitative results and discussion, this section presents an
end-to-end case study using a flange-type component with a central threaded hub,
a central bore, a six-hole bolt pattern, and edge-finishing features. The case
study illustrates how the proposed framework transforms a realistic pair of
design artifacts, namely a 3D CAD model and a 2D engineering drawing, into a
fused feature representation and a traceable manufacturing process plan. It also
provides a concrete view of intermediate outputs, context-fusion behavior, HITL
correction, and final planning decisions.

As shown in Fig. 12, the case begins with two design inputs: the 3D CAD model of
the flange and its corresponding engineering drawing. The Feature Extraction
Agent processes these inputs through parallel 3D and 2D branches and returns
aggregated intermediate results. From the 3D CAD model, the system identifies
the main manufacturable features, including the outer flange body, the six-hole
pattern, the central hub region, the external thread region, the central bore,
and edge-finishing features such as the chamfer and fillet. In parallel, the 2D
drawing analysis pipeline extracts the corresponding engineering specifications,
including the outer diameter, bolt-circle diameter, six equally spaced holes,
the M42$\times$1.5-6g thread callout, the central bore dimension, the chamfer
specification, the fillet specification, and the associated GD\&T and datum
references. These outputs are consolidated into a structured intermediate
representation that preserves both geometric feature information and
drawing-derived design intent for downstream reasoning.

\begin{figure}[!htbp]
\centering
\includegraphics[width=\textwidth,height=0.84\textheight,keepaspectratio]{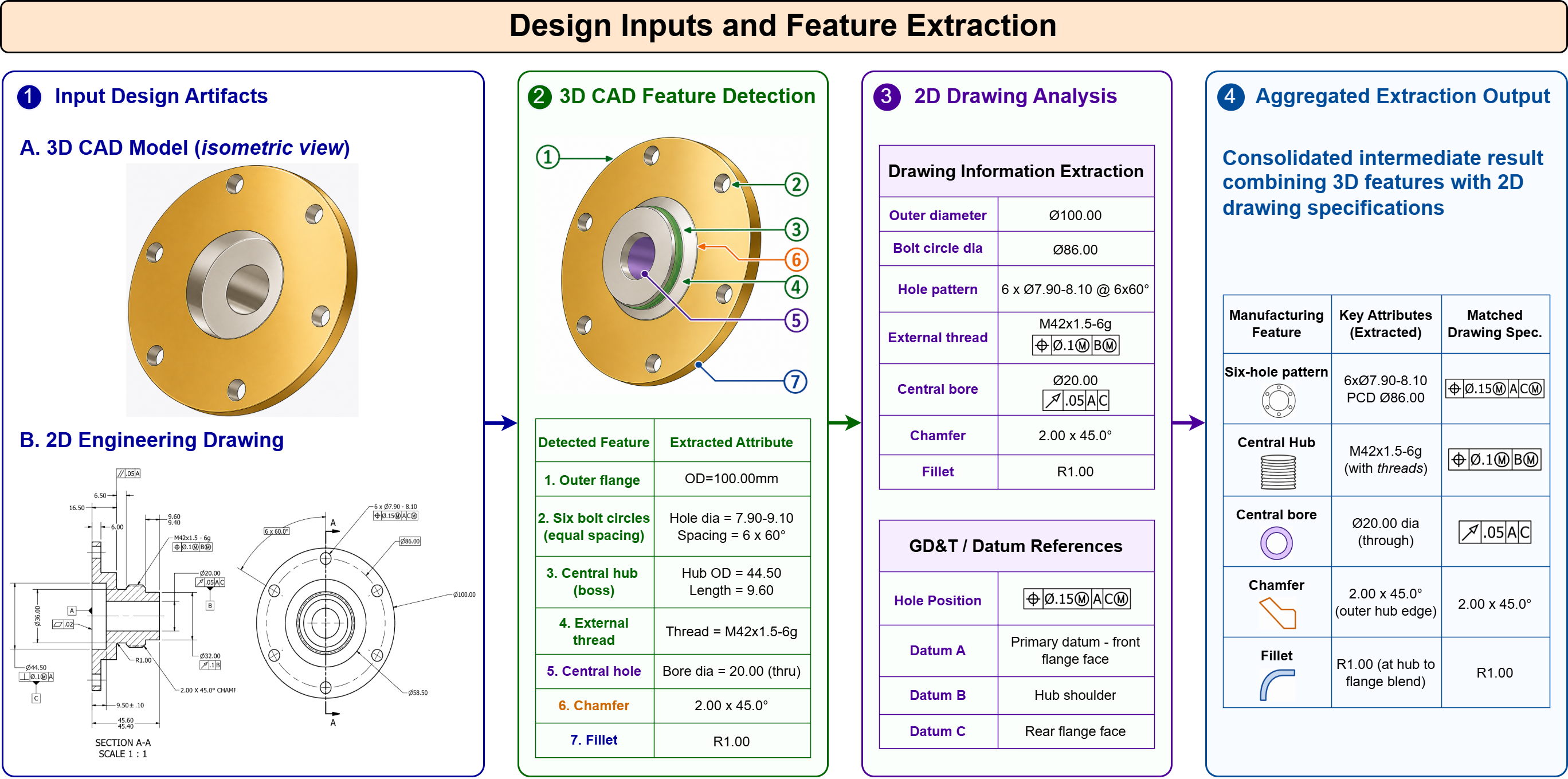}
\caption{Case study feature extraction workflow, showing how the flange CAD
model and engineering drawing are processed to identify 3D manufacturing
features and extract 2D drawing specifications.}
\end{figure}

The role of the Context Fusion Agent is illustrated in Fig. 13. This stage is
essential because the extracted 3D features and 2D annotations are not directly
actionable unless they are linked at the feature level. The fusion process
begins with semantic interpretation, where the 2D drawing callouts are converted
into structured semantic tags such as hole-pattern size, bolt-circle location,
external thread specification, central bore specification, edge chamfer, and
datum or tolerance context. The agent then performs hybrid matching by jointly
considering geometric cues, spatial context, and semantic type to associate each
annotation with the most plausible 3D feature. In this case, most mappings are
resolved automatically. For example, the six-hole pattern callout is mapped to
the detected bolt-hole pattern, the central-bore callout is linked to the coaxial
inner bore, and the 2.00 $\times$ 45\textdegree{} callout is mapped to the edge
chamfer. However, one ambiguity is observed for the M42$\times$1.5-6g
specification because the coaxial arrangement of the inner bore and the external
hub creates a potential mismatch during automatic association. The initial
ambiguous mapping is corrected during the HITL review stage by reassigning the
thread callout to the external hub. After this correction, the final fused
representation contains validated feature-level manufacturing context and is
forwarded to the downstream analytical agents.

\begin{figure}[!htbp]
\centering
\includegraphics[width=\textwidth,height=0.84\textheight,keepaspectratio]{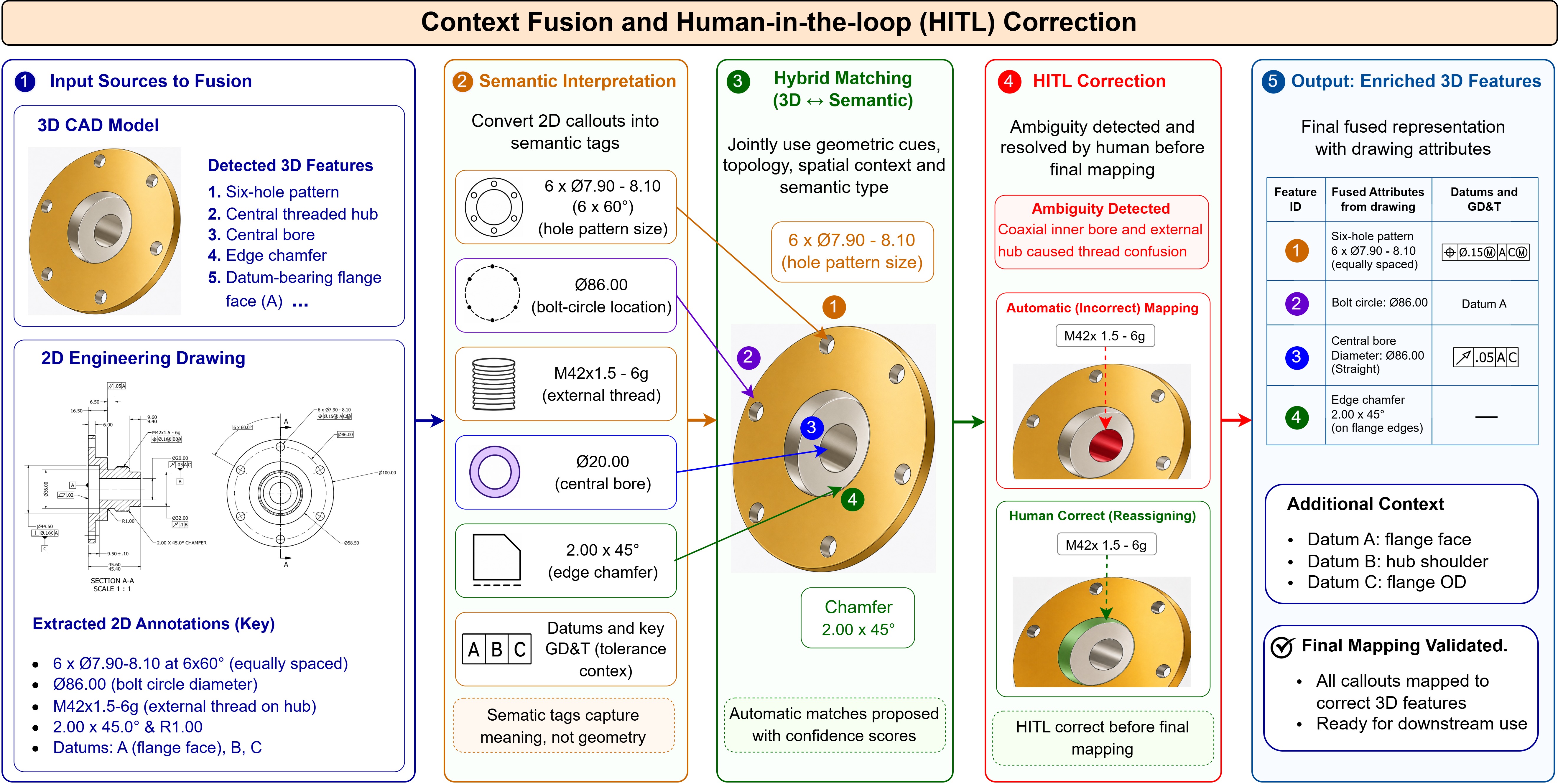}
\caption{Case study context fusion workflow, showing semantic annotation
tagging, 2D--3D feature matching, and HITL correction of an ambiguous
M42$\times$1.5-6g thread mapping.}
\end{figure}

The downstream planning results are shown in Fig. 14. Based on the fused feature
representation, the KR Agent retrieves manufacturing knowledge relevant to the
hole pattern, external thread, central bore, chamfer, and datum structure. The
retrieved knowledge includes machining guidance for the six-hole pattern under
positional tolerance control, threading guidance for the M42$\times$1.5-6g
external thread, setup guidance for maintaining concentricity between the hub
and bore features, and finishing considerations for the chamfer and fillet.
Using this information, the PS Agent generates an example operation plan
consisting of facing and datum establishment, turning of the outer flange and
hub region, drilling and boring of the central hole, external threading of the
hub, drilling of the six-hole bolt pattern on the specified pitch-circle
diameter, edge finishing for the chamfer and fillet features, and final
inspection. The TS Agent then assigns representative tooling, including a facing
and OD-turning tool, drilling and boring tools, an external threading tool, a
drill for the hole pattern, and a chamfering tool. Inspection resources are also
recommended, including a GO/NO-GO thread gauge for thread verification and
position-verification resources for the hole pattern.

\begin{figure}[!htbp]
\centering
\includegraphics[width=\textwidth,height=0.84\textheight,keepaspectratio]{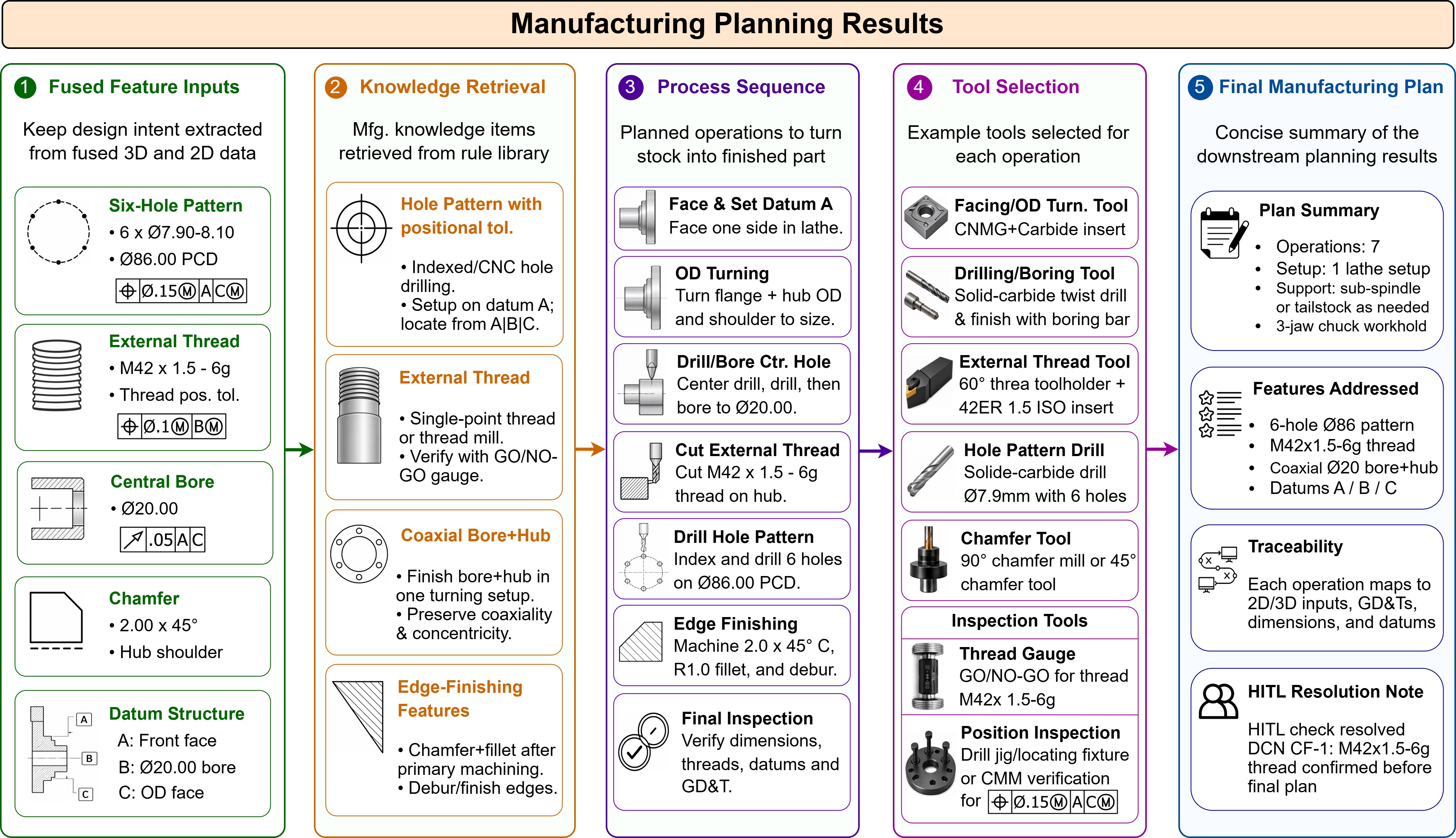}
\caption{Case study manufacturing planning workflow, showing how the fused
feature representation supports knowledge retrieval, process sequencing, tool
selection, and final plan generation for the flange component.}
\end{figure}

This case study demonstrates how the proposed framework maintains a continuous
information flow from realistic design inputs to manufacturing planning outputs.
The example shows that the system can preserve both CAD-derived geometric
features and drawing-derived engineering specifications throughout the workflow.
It also highlights that context fusion is a core reasoning stage rather than a
simple matching operation, because feature-level links are required before
annotations such as tolerances, threads, surface finish requirements, and datum
references can support manufacturing decisions. The HITL correction further
illustrates how expert feedback can resolve ambiguous associations when
geometrically related features create uncertainty. Overall, the case study
complements the quantitative evaluation by showing how the proposed multi-agent
framework produces a traceable manufacturing plan in which retrieved knowledge,
process decisions, tool recommendations, and inspection considerations remain
linked to the original design intent.

\HeadingOne{6. Conclusions}

This paper presented Design-to-Plan, a fully implemented agentic framework for
manufacturing process planning from heterogeneous design artifacts. The
framework connected 3D CAD feature recognition, 2D engineering drawing analysis,
context fusion, knowledge retrieval, process sequencing, tool selection, and
report generation within a coordinated multi-agent workflow. Its primary
contribution is not simply the use of LLMs in manufacturing, but the deployment
of LLMs as interactive reasoning agents that coordinate with deterministic
modules, manufacturing knowledge sources, and specialized agents to produce
traceable planning outputs.

The proposed hybrid deterministic-agentic design assigns each task to the most
suitable computational paradigm. Deterministic modules support reliable
extraction of CAD and drawing information, while LLM-based agents perform
tool-grounded reasoning over incomplete, ambiguous, and potentially conflicting
manufacturing information. A central component of this workflow is 2D-3D context
fusion, which links CAD-derived manufacturing features with drawing-derived
specifications such as dimensions, tolerances, GD\&T annotations, surface finish
requirements, material information, and manufacturing notes. This fused
representation provides the feature-level manufacturing context required for
downstream planning decisions.

The evaluation shows that the framework can reliably coordinate multiple
specialized agents across diverse input conditions. The sequential architecture
provides stronger performance in quality-oriented reasoning, while the parallel
architecture improves robustness, coverage, and computational efficiency. These
results demonstrate a practical quality-efficiency trade-off between deeper
validation and scalable execution. The case study further shows that the
proposed workflow can maintain traceability from original design inputs to final
manufacturing planning outputs.

Overall, Design-to-Plan demonstrates one of the first end-to-end agentic
frameworks for transforming CAD models and engineering drawings into executable
manufacturing process plans. By linking design interpretation, manufacturing
knowledge, process decisions, tooling, and reporting, the framework provides a
practical step toward intelligent design-to-manufacturing automation. Future
work will extend this foundation toward broader agentic manufacturing
intelligence, including CAM strategy generation, machining parameter
optimization, cost-time-quality trade-off analysis, process monitoring, and
inspection-driven feedback. From the agentic AI perspective, future development
will move beyond prompt-level design toward more reliable harness engineering,
where agents are supported by richer context, validated tools, reusable skills,
and closed-loop verification mechanisms for robust industrial deployment.

\HeadingOne{Declaration of Competing Interest}

The authors declare that they have no known competing financial interests or
personal relationships that could have appeared to influence the work reported
in this paper.

\HeadingOne{Acknowledgements}

This work is supported by Singapore International Graduate Award (SINGA)
(Awardee: Muhammad Tayyab Khan) funded by Agency for Science, Technology and
Research (A*STAR) and Nanyang Technological University, Singapore.

\HeadingOne{References}

\begin{spacing}{1.0}
\bibliographystyle{unsrtnat}
\bibliography{references}
\end{spacing}

\clearpage
\setcounter{figure}{0}
\renewcommand{\thefigure}{A\arabic{figure}}

\HeadingOne{Appendix. Knowledge Retrieval Agent: Parallel ReAct System Prompts}

The parallel ReAct architecture described in Section 3.4.2 partitions the
knowledge tools across three specialized sub-agents that operate concurrently,
followed by a coordinator that merges their outputs using the same
source-priority hierarchy as the sequential configuration. The structured-data
sub-agent handles SQL, tabular, and decision-tree tools. The text-knowledge
sub-agent processes RAG-based text and knowledge-graph queries. The material
sub-agent retrieves material-specific constraints. Each sub-agent prompt defines
a restricted tool scope while enforcing consistent provenance requirements. The
coordinator prompt applies conflict resolution rules and integrates results into
a unified response.

\begin{figure}[!htbp]
\centering
\includegraphics[width=0.849\textwidth,height=0.84\textheight,keepaspectratio]{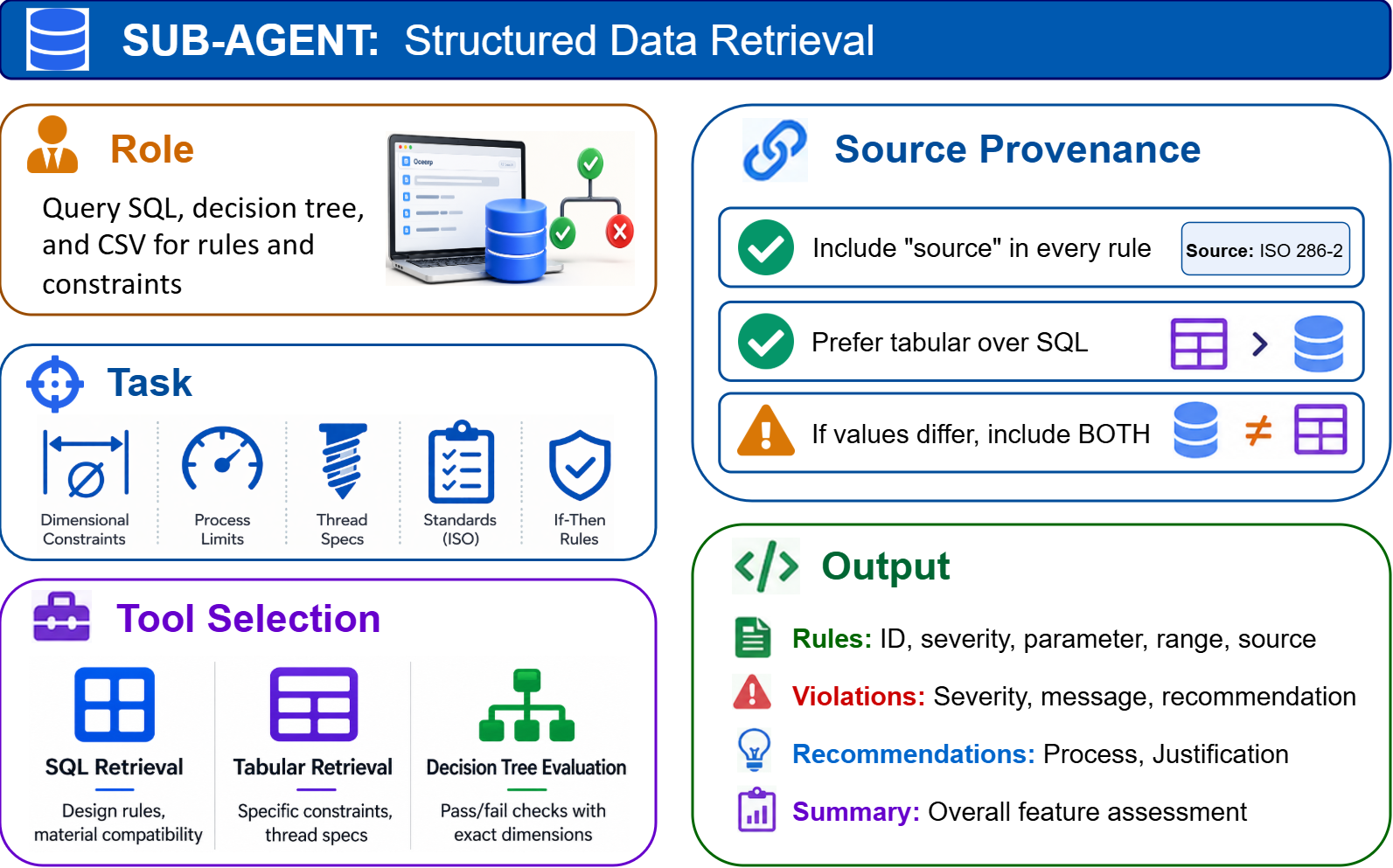}
\caption{Structured-data sub-agent prompt. Handles SQL, tabular, and rule-based
queries for quantitative constraints; conflicting outputs are forwarded for
resolution.}
\end{figure}

\begin{figure}[!htbp]
\centering
\includegraphics[width=0.852\textwidth,height=0.84\textheight,keepaspectratio]{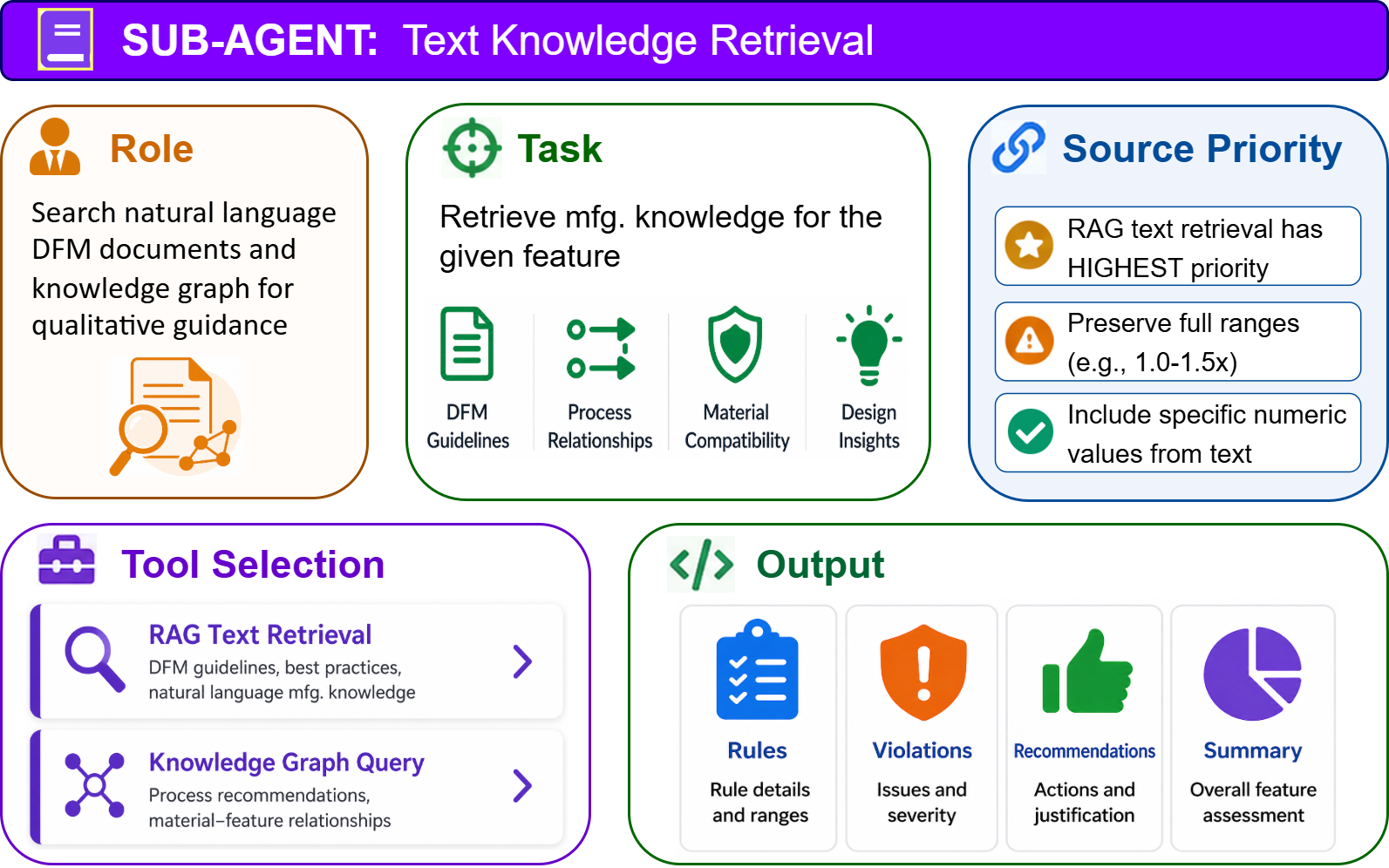}
\caption{Text-knowledge sub-agent prompt. Retrieves qualitative guidelines from
RAG text and knowledge graphs while preserving source provenance.}
\end{figure}

\begin{figure}[!htbp]
\centering
\includegraphics[width=0.895\textwidth,height=0.84\textheight,keepaspectratio]{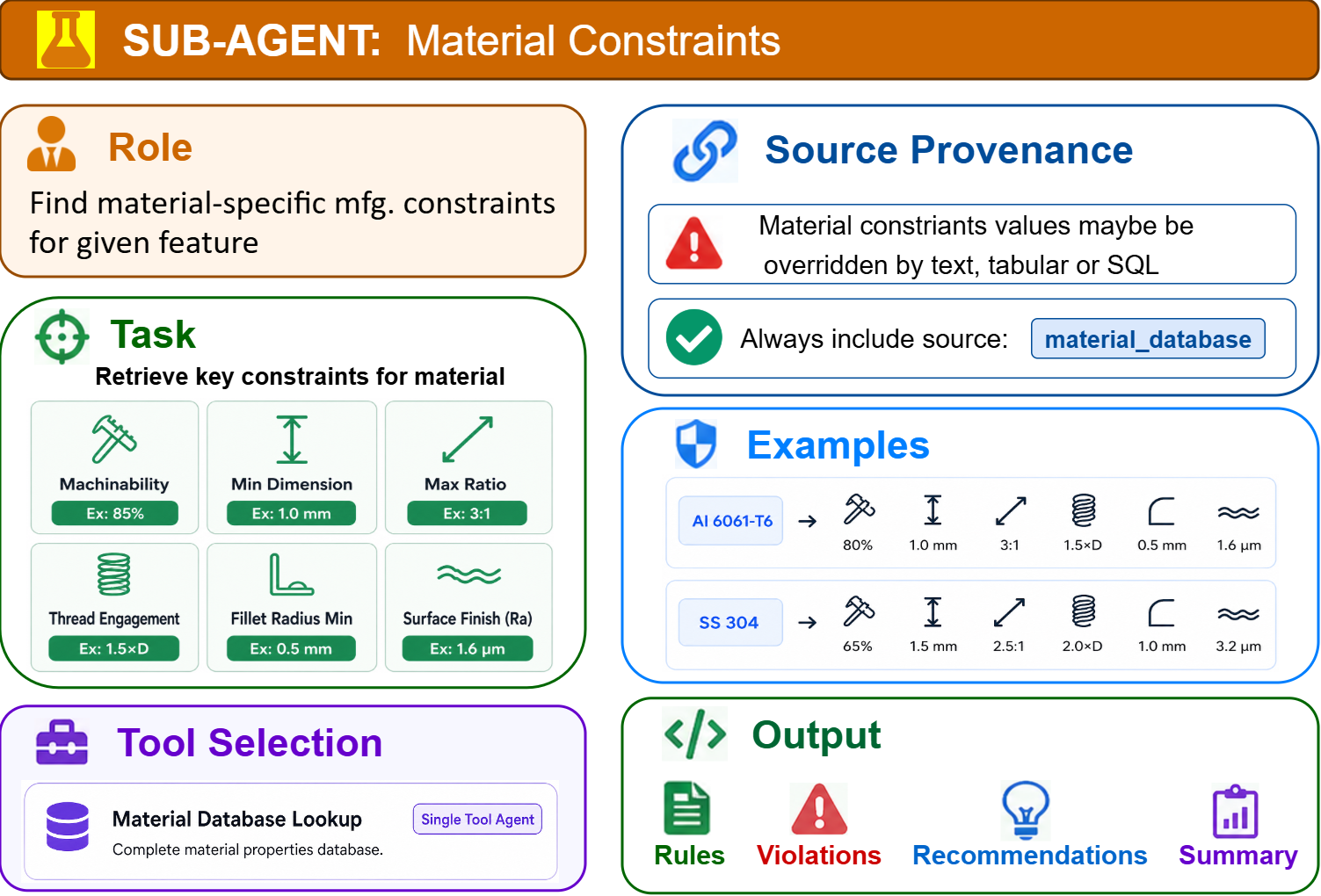}
\caption{Material sub-agent prompt. Retrieves material-specific constraints such
as machinability and dimensional limits.}
\end{figure}

\begin{figure}[!htbp]
\centering
\includegraphics[width=0.933\textwidth,height=0.84\textheight,keepaspectratio]{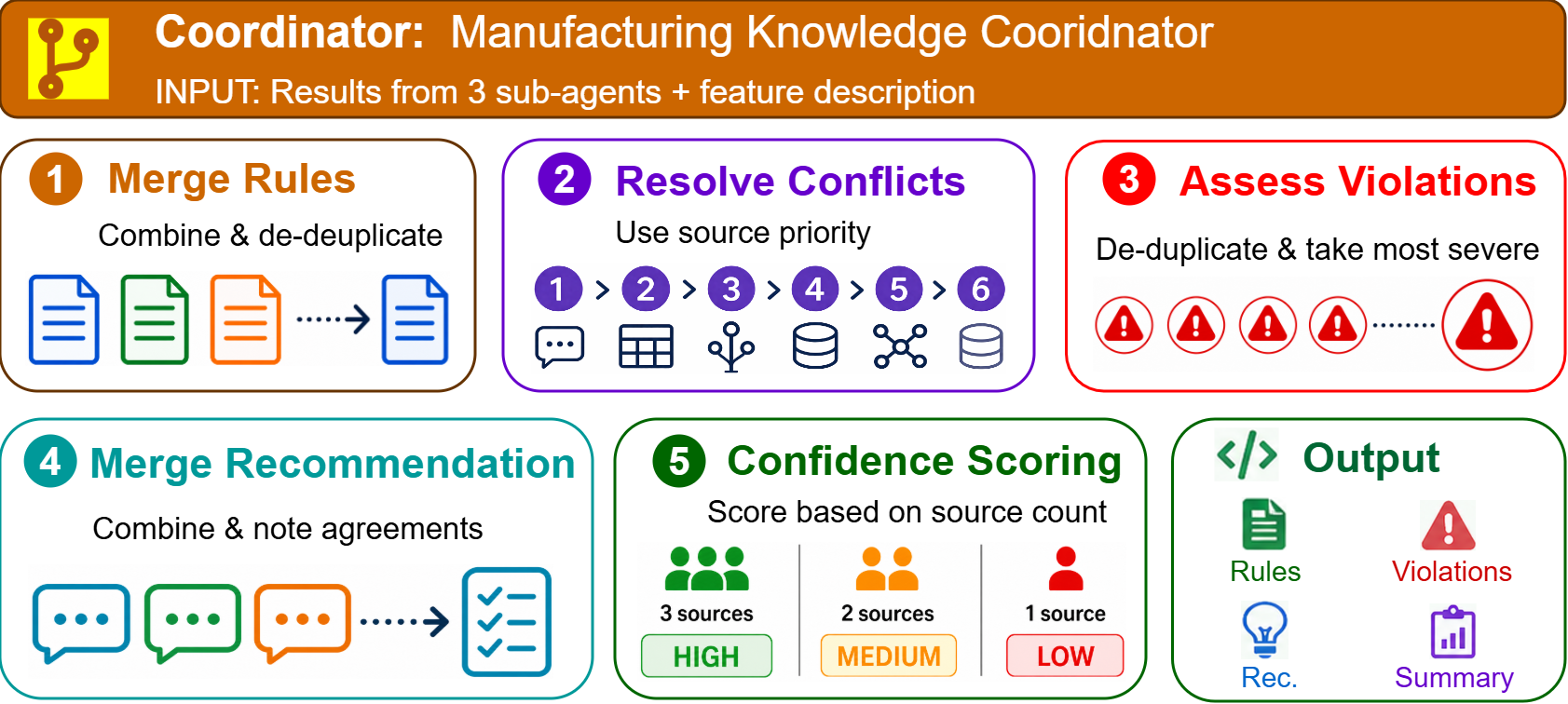}
\caption{Coordinator prompt. Merges sub-agent outputs, resolves conflicts using
the source-priority hierarchy, and produces a unified response with provenance
and confidence scores.}
\end{figure}

\end{document}